\documentclass{article}

\usepackage{microtype}
\usepackage{graphicx}
\usepackage{subfigure}
\usepackage{booktabs} 

\usepackage{hyperref}

\usepackage[accepted]{icml2023}

\usepackage{amsmath}
\usepackage{amssymb}
\usepackage{mathtools}
\usepackage{amsthm}

\usepackage[capitalize,noabbrev]{cleveref}

\theoremstyle{plain}

\theoremstyle{definition}

\theoremstyle{remark}

\usepackage[textsize=tiny]{todonotes}

\usepackage[utf8]{inputenc} 
\usepackage[T1]{fontenc}    
\usepackage{hyperref}       
\usepackage{url}            
\usepackage{booktabs}       
\usepackage{amsfonts}       
\usepackage{nicefrac}       
\usepackage{microtype}      
\usepackage{xcolor}         
\usepackage{times}
\usepackage{epsfig}
\usepackage{graphicx}
\usepackage{amsmath}
\usepackage{amssymb}
\usepackage{dsfont}
\usepackage{multirow}
\usepackage{enumitem}
\usepackage{pifont}
\usepackage{comment}
\usepackage{caption}
\usepackage{silence}
\usepackage{physics}
\usepackage{nicematrix}
\usepackage{wrapfig}
\usepackage{float}
\usepackage{tikz}
\usepackage{algpseudocode}

\newcommand{\cmark}{\ding{52}}%
\newcommand\calA{\mathcal{A}} 

\newcommand\calM{\mathcal{M}}

\usepackage{amsmath,amsfonts,bm}

\def\eqref#1{equation~\ref{#1}}

\def\1{\bm{1}}

\DeclareMathAlphabet{\mathsfit}{\encodingdefault}{\sfdefault}{m}{sl}
\SetMathAlphabet{\mathsfit}{bold}{\encodingdefault}{\sfdefault}{bx}{n}

\newcommand{\sigmoid}{\sigma}

\def\halfcheckmark{\textcolor{black}{\ding{51}}{\small\textcolor{black}{\kern-0.7em\ding{55}}}}

\icmltitlerunning{Open-Vocabulary Universal Image Segmentation with MaskCLIP}

\begin{document}

\twocolumn[
\icmltitle{Open-Vocabulary Universal Image Segmentation with MaskCLIP}



\icmlsetsymbol{equal}{*}

\begin{icmlauthorlist}
\icmlauthor{Zheng Ding}{ucsd}
\icmlauthor{Jieke Wang}{ucsd}
\icmlauthor{Zhuowen Tu}{ucsd}
\end{icmlauthorlist}

\icmlaffiliation{ucsd}{University of California San Diego, Lo Jolla, CA 92093, USA}

\icmlcorrespondingauthor{Zheng Ding}{zhding@ucsd.edu}
\icmlcorrespondingauthor{Jieke Wang}{jiw010@ucsd.edu}
\icmlcorrespondingauthor{Zhuowen Tu}{ztu@ucsd.edu}

\icmlkeywords{Machine Learning, ICML}

\vskip 0.3in
]



\printAffiliationsAndNotice{}  

\begin{abstract}
In this paper, we tackle an emerging computer vision task, open-vocabulary universal image segmentation, that aims to perform semantic/instance/panoptic segmentation (background semantic labeling + foreground instance segmentation) for arbitrary categories of text-based descriptions in inference time. 
We first build a baseline method by directly adopting pre-trained CLIP models without finetuning or distillation. We then develop MaskCLIP, a Transformer-based approach with a MaskCLIP Visual Encoder, which is an encoder-only module that seamlessly integrates mask tokens with a pre-trained ViT CLIP model for semantic/instance segmentation and class prediction. MaskCLIP learns to efficiently and effectively utilize pre-trained partial/dense CLIP features within the MaskCLIP Visual Encoder that avoids the time-consuming student-teacher training process. MaskCLIP outperforms previous methods for semantic/instance/panoptic segmentation on ADE20K and PASCAL datasets. We show qualitative illustrations for MaskCLIP with online custom categories. Project website: \href{https://maskclip.github.io}{https://maskclip.github.io}.

\end{abstract}

\section{Introduction}

Panoptic segmentation  \citep{kirillov2018panoptic} or image parsing  \citep{tu2005image} integrates the task of semantic segmentation  \citep{tu2008auto} for background regions (e.g. ``stuff'' like ``road'', ``sky'') and instance segmentation  \citep{he2017mask} for foreground objects (e.g. ``things'' such as ``person'', ``table''). Existing panoptic segmentation methods  \citep{kirillov2018panoptic,kirillov2019panoptic,li2018attention,xiong2019upsnet,lazarow2020learning} and instance segmentation approach  \citep{he2017mask} deal with a fixed set of category definitions, which are essentially represented by categorical labels without semantic relations. DETR  \citep{carion2020end} is a pioneering work that builds a Transformer-based architecture for both object detection and panoptic segmentation. Under a more general setting, the tasks of semantic \citep{tu2008auto}, instance \citep{he2017mask}, and panoptic \citep{kirillov2018panoptic} can be unified under a universal image segmentation paradigm \citep{cheng2021masked}. 

The deep learning field is moving rapidly towards the open-world/zero-shot settings  \citep{bendale2015towards} where computer vision tasks such as classification \citep{CLIP}, object detection \citep{li2021grounded,zareian2021open,zang2022open,gu2022open,cai2022x}, semantic labeling \citep{li2022language,ghiasi2021open}, and image retrieval  \citep{bendale2015towards,hinami2017discriminative,zareian2021open,hinami2017discriminative,kamath2021mdetr} perform recognition and detection for categories beyond those in the training set.

In this paper, we take advantage of the existence of pre-trained CLIP image and text embedding models \citep{CLIP}, that are mapped to the same space. We first build a baseline method for open-vocabulary panoptic segmentation using CLIP models without training. We then develop a new algorithm, MaskCLIP, that is a Transformer-based approach efficiently and effectively utilizing pre-trained partial/dense CLIP features without heavy re-training. The key component of MaskCLIP is a Relative Mask Attention (RMA) module that seamlessly integrates the mask tokens  with a pre-trained ViT-based CLIP backbone. MaskCLIP is distinct and advantageous compared with previous approaches in three aspects: 
1) A canonical background and instance segmentation representation by the mask token representation with a unique encoder-only strategy that tightly couples a pre-trained CLIP image feature encoder with the mask token encoder.
2) MaskCLIP avoids the challenging student-teacher distillation processes such as OVR-CNN  \citep{zareian2021open} and ViLD  \citep{gu2022open} that face a limited number of teacher objects to train; 3) MaskCLIP also learns to refine masks beyond simple pooling in e.g. OpenSeg  \citep{ghiasi2021open}.

The contributions of our work are listed as follows.
\begin{itemize}[leftmargin=0.5cm]
    \vspace{-2mm}
     \item We develop a new algorithm, MaskCLIP, to perform open-vocabulary universal image segmentation building on top of canonical background and instance mask representation with a cascade mask proposal and refinement process.
    \item We device the MaskCLIP Visual Encoder under an encoder-only strategy by tightly coupling a pre-trained CLIP image feature encoder with the mask token encoder, to allow for the direct formulation of the mask feature representation for semantic/instance segmentation+refinement, and class prediction. Within the MaskCLIP Visual Encoder, there is a new module called Relative Mask Attention (RMA) that performs mask refinement. 
   
    \item  MaskCLIP expands the scope of the CLIP models to open-vocabulary universal image segmentation by demonstrating encouraging and competitive results for open-vocabulary semantic, instance, and panoptic segmentation.
\end{itemize}

\begin{table*}[!htb]
\centering
\caption{\small Comparison for recent open-vocabulary approaches for object detection, semantic segmentation, instance segmentation, and panoptic segmentation. GLIP \citep{li2021grounded}; OVR-CNN \citep{zareian2021open}; ViLD \citep{gu2022open}; RegionCLIP \citep{zhong2021regionclip}; OV-DETR \citep{zang2022open}; LSeg \citep{li2022language}; OPenSeg \citep{ghiasi2021open}; DenseCLIP \citep{rao2022denseclip}; XPM \citep{huynh2022open}. \halfcheckmark \; indicates that the corresponding method is loosely following the definition. Dense Clip features refer to the use of pixel-wise/local features. Note that OpenSeg uses its ALIGN \citep{jia2021scaling}, which is an alternative to CLIP.}\label{tab:related_comparison}
\scalebox{0.68}{%
\begin{tabular}{cccccccc}
Task & Method & Arbitrary Online & \multicolumn{2}{c}{Segmentation} & Dense CLIP & Training & Annotation \\
  \cmidrule(lr){4-5}
    & & Inference & semantic & instance & features & data & type \\
\toprule
\multirow{4}{*} {Object Det.} & GLIP & \cmark &  &  & & {\scriptsize FourODs, GoldG, Cap24M} & labels + bbox + captions \\
& OVR-CNN & \cmark &  &  & & {\scriptsize COCO base, CC3M} & bbox + captions \\
& ViLD & \cmark &  &  & & {\scriptsize COCO} & labels + bbox \\
 & RegionCLIP & \cmark &  &  &  & {\scriptsize CC3M, COCO} & captions \\
\midrule
\multirow{3}{*} {Semantic Seg.} & LSeg  & \halfcheckmark & \cmark &  & & {\scriptsize COCO + others} 
& labels + segmentations \\
 & OpenSeg & \cmark & \cmark & \halfcheckmark & \cmark & {\scriptsize COCO, LocalizedNarratives} & masks + captions \\
& DenseCLIP &  & \cmark &  & \cmark & {\scriptsize COCO} & labels + segmentations \\
\midrule
\multirow{1}{*} {Instance Seg.} & XPM & \halfcheckmark &  & \cmark & & {\scriptsize COCO, CC3M} & labels + masks + captions \\ 
\midrule
\multirow{1}{*} {Panoptic Seg.} & MaskCLIP (ours) & \cmark & \cmark & \cmark & \cmark & {\scriptsize COCO} & labels + masks \\
\bottomrule
\end{tabular}
}
\end{table*}

\section{Related Work}

\textbf{Open vocabulary.} The open vocabulary setting is gaining increasing popularity lately as traditional fully supervised settings cannot handle unseen classes during testing, while real-world vision applications like scene understanding, self-driving and robotics are commonly required to predict unseen classes. Previous open-vocabulary attempts have been primarily made for object detection.
ViLD \citep{gu2022open} trains a student model to distill the knowledge of CLIP. RegionCLIP \citep{zhong2021regionclip} finetunes the pretrained CLIP model to match the image areas with corresponding texts. OV-DETR  \citep{zang2022open} uses CLIP as an external model to obtain the query embedding from CLIP model. Recently there is also work made for open-vocabulary semantic segmentation \citep{ghiasi2021open}. 

\textbf{Universal segmentation.} 
Previously semantic/instance/panoptic segmentation tasks have been treated as different tasks using different methods. With the recent trends in computer vision, the formulation and methods of the three segmentation tasks have gradually been uniformed \citep{cheng2021per, cheng2021masked}. Instead of separately dealing with the stuff/instance, those methods treat them as the same one and output masks for each stuff/instance and do a post-process on the output masks for different segmentation tasks.

\textbf{Open-vocabulary universal segmentation: an emerging task.} As open-set, open-world, zero-shot, and open-vocabulary are relatively new concepts that have no commonly accepted definitions, thus, different algorithms are often not directly comparable with differences in problem definition/setting, training data, and testing scope. Table \ref{tab:related_comparison} gives a summary for the recent open-vocabulary applications.
XPM \citep{huynh2022open} utilizes vision-language cross-modal data to generate pseudo-mask supervision to train a student model for instance segmentation, and thus, it may not be fully open-vocabulary to allow for arbitrary object specifications in the inference time. LSeg  \citep{li2022language} also has a limited open-vocabulary aspect as the learned CNN image features in LSeg  are not exposed to representations beyond the training labeling categories. OpenSeg \citep{ghiasi2021open} is potentially applicable for instance/panoptic segmentation, but OpenSeg is formulated to be trained on captions that lack instance-level information that is fundamental for panoptic segmentation.
The direct image feature pooling strategy in OpenSeg is potentially another limiting factor towards the open-vocabulary universal segmentation.
Nevertheless, no results for open-vocabulary panoptic/instance segmentation are reported in  \citep{ghiasi2021open}.  

\textbf{Class-agnostic segmentation.} Most closed-vocabulary segmentation methods are class-ware i.e. predicting the classes along with the corresponding labels \citep{he2017mask, cheng2021per, cheng2021masked}. However, in tasks involving open-vocabulary or open-world scenarios where novel classes may appear during testing, it is common to use class-agnostic segmentation methods for generating masks\cite{jia2021scaling, qi2022open, xu2022simple}. The difference in methodology between class-aware and class-agnostic segmentation methods is typically not substantial. Class-aware methods often incorporate a class-prediction head, whereas class-agnostic methods do not. In our method, we adopt a class-agnostic segmentation model by removing the class-prediction head from previous class-aware class-aware segmentation methods.

\textbf{CLIP model distillation/reuse.} After its initial release, the CLIP model  \citep{CLIP} that is learned from large-scale image-text paired captioning datasets has received a tremendous amount of attention. Some other similar vision-language models have also been proposed later e.g. ALIGN \citep{jia2021scaling}, GLIP \citep{li2021grounded}.  Many algorithms have been developed lately  \citep{zang2022open,wang2022clip,zhong2021regionclip,luo2021clip4clip,patashnik2021styleclip,shen2021much} trying knowledge distillation from the CLIP model to benefit the down-stream tasks one way or the other by leveraging the rich semantic language information paired in the images. Here, we directly adopt the backbone of CLIP image model to train for open-vocabulary panoptic segmentation. There have been attempts \citep{rao2022denseclip,zhou2022extract} that use the partial/dense CLIP features to represent pixel-wise features as teacher model to train student model for semantic segmentation. 

\section{Method}

\begin{figure*}[h]
    \centering
    \includegraphics[width=0.9\textwidth]{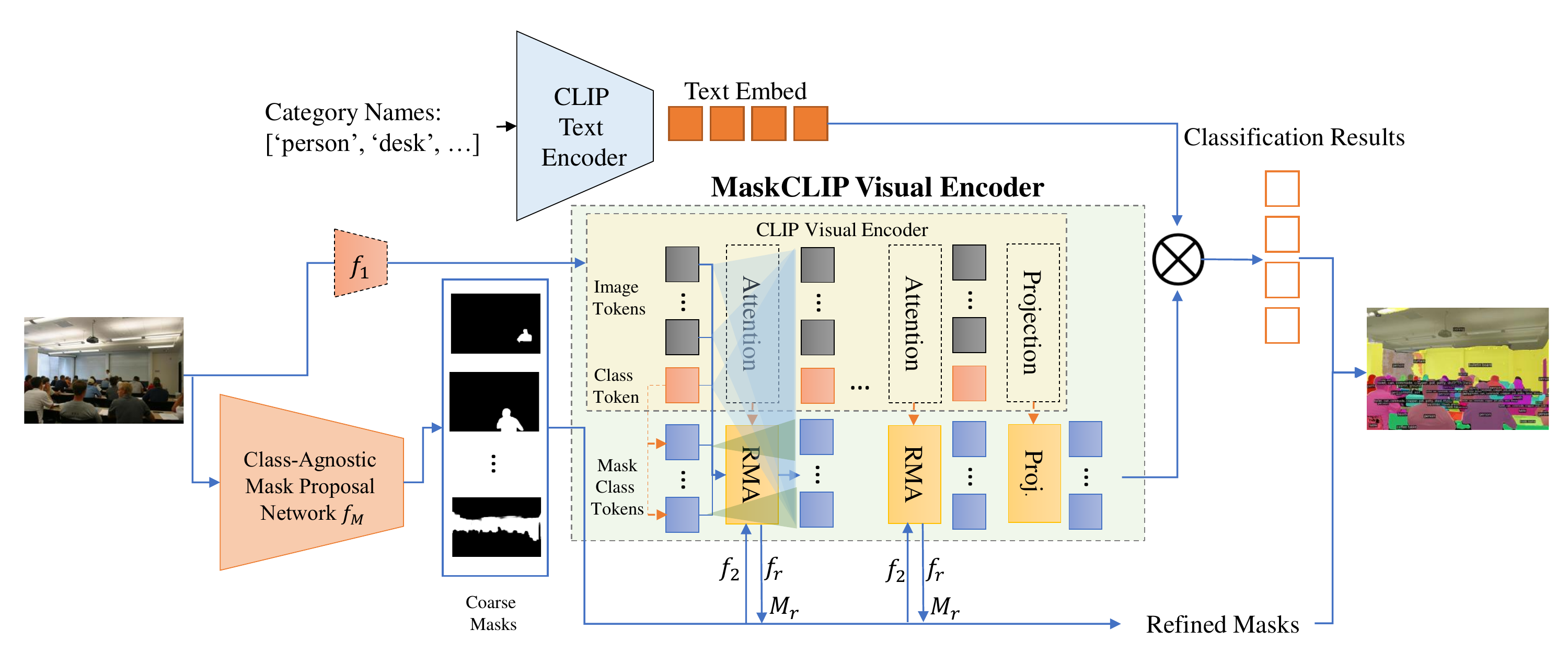}
    \caption{\small \textbf{Illustration of the pipeline.} Our pipeline contains two stages. The first stage is a class-agnostic mask proposal network and the second stage is built on the pretrained CLIP ViT model. All the weights of the CLIP ViT model during training are fixed. Arrows in orange denote weight sharing. The embeddings' weights of Mask Class Tokens are shared by Class Tokens in the CLIP ViT model and are fixed. RMA represents Relative Mask Attention which is built based on the CLIP ViT attention layer. RMA contains all the weights from CLIP ViT attention layer which are all fixed during training. Additional weights are added in RMA for further mask information utilization and mask refinement. The demo image we use here is from ADE20K \citep{zhou2016semantic}.}
    \label{fig:network}
\end{figure*}

Our pipeline, shown in Figure \ref{fig:network}, contains two stages. The first stage is a class-agnostic mask proposal network. The second stage is MaskCLIP Visual Encoder which is built on the CLIP \citep{CLIP} ViT architecture. It takes the images and the coarse masks from the first stage as the input and outputs refined masks along with the corresponding partial/dense image features for further classification.
{
\subsection{Class-Agnostic Mask Proposal Network}

Our Class-Agnostic Mask Proposal Network is built on instance/segmentation models such as MaskRCNN\citep{he2017mask} and Mask2Former\citep{cheng2021masked}. To make the model class-agnostic, we remove the class supervision during training. The classification head thus becomes a binary classification for either positive or negative in these models.

\subsection{MaskCLIP Visual Encoder}

Similar to CLIP, our MaskCLIP Visual Encoder also predicts the image features. Unlike the CLIP Visual Encoder which only uses one class token to output the feature of the whole image. Our MaskCLIP Visual Encoder uses another $M$ Mask Class Tokens to output the partial/dense features for each corresponding area of the image given the masks. The Mask Class Tokens use attention masks and Relative Mask Attention to obtain the partial/dense features which we discuss in the following two parts.
}
\label{sec:masktokens}
\paragraph{Mask Class Tokens.} In order to obtain partial/dense image features for the corresponding masks or bounding boxes for further recognition or distillation, an easy way to do this is simply masking or cropping the image and then sending the obtained image to the pretrained image encoder. This method has been widely used in several open vocabulary detection/segmentation methods \citep{zhong2021regionclip, gu2022open, xu2022simple}. The problem is that it's not computation efficient ($N$ masks/boxes will lead to $N$ images and they will be computed through the image encoder independently) and also loses the ability to see the global image context information which is very important for recognizing some objects and stuff. For masking, another problem is that masks are in different shapes and simply masking the image will cause the resulting image to have a transparent background which usually doesn't exist in real images that are used for training in large language-vision models e.g., CLIP.

To solve this, we propose Mask Class Tokens for efficient feature extraction from images without losing the global image context information. In the original CLIP ViT-based visual encoder framework, the input of the network is $N$ image tokens and $1$ class token. The final output of the class token will be used for the relation computation with the text embeddings. Our newly introduced $M$ Mask Class Tokens will be alongside the image tokens and the class token. The embeddings' weights of the Mask Class Token are provided by the class token in the pretrained CLIP ViT model and are fixed. Each Mask Class Token will output a corresponding partial/dense image feature similar to the class token which outputs the feature of the whole image. To achieve this, we design an attention mask as follows
\begin{align}
    \calM = \begin{bmatrix}
\multicolumn{2}{c}{\mathcal{F}_{(N+1)\times (N+1)}} & \mathcal{T}_{(N+1)\times M}\\
 \mathcal{M}^{'}_{M\times N} & \mathcal{F}_{M\times 1} & \mathcal{T}_{M\times M}
\end{bmatrix}
\label{eq:attn_mask}
\end{align}
in which $M$ is the number of Mask Class Tokens, $N$ is the number of image tokens, $\mathcal{T}_{m\times n}$ is an $m\times n$ True matrix, $\mathcal{F}_{m\times n}$ is an $m\times n$ False matrix and $\calM'$ is defined as following:
\begin{align}
    \calM^\prime_{i,j} =     \begin{cases}
      \text{False} & \text{if mask$_{i}$ contains at least one pixel in patch$_{j}$}\\
      \text{True} & \text{otherwise.}
    \end{cases}
\end{align}
where True means that this position is masked out i.e. not allowed to attend and False otherwise. 

 In our mask attention matrix $\calM$, $\mathcal{F}_{(N+1)\times (N+1)}$ shows the $N$ Image Tokens and one Class Token are attending each other as in the original CLIP. $\mathcal{T}_{(N+1)\times M}$ shows that the $N$ Image Tokens and one Class Token are not attending the $M$ Mask Class Tokens. $\mathcal{M}^{'}_{M\times N}$ shows that the Mask Class Tokens are attending the Image Tokens given the corresponding masks. $\mathcal{F}_{M\times 1}$ shows that the $M$ Mask Class Tokens are attending the Class Token. $\mathcal{T}_{M\times M}$ shows that the $M$ Mask Class Tokens are not interacting with each other.

In this way, each Mask Class Token will learn from the corresponding mask area of the images. The image tokens are also interacting with each other which means the global information won't lose. And it's also very efficient since we don't need to do redundant computing for each mask or finetune the pretrained model. However, the mask information is not fully utilized and it cannot be refined either. But we will see in the experiments later that simply adopting Mask Class Tokens to the pretrained CLIP model without any finetuning will already serve as a competitive baseline.

\label{sec:rma}

\begin{figure}[!htp]
\begin{center}
\includegraphics[width=0.9\linewidth]{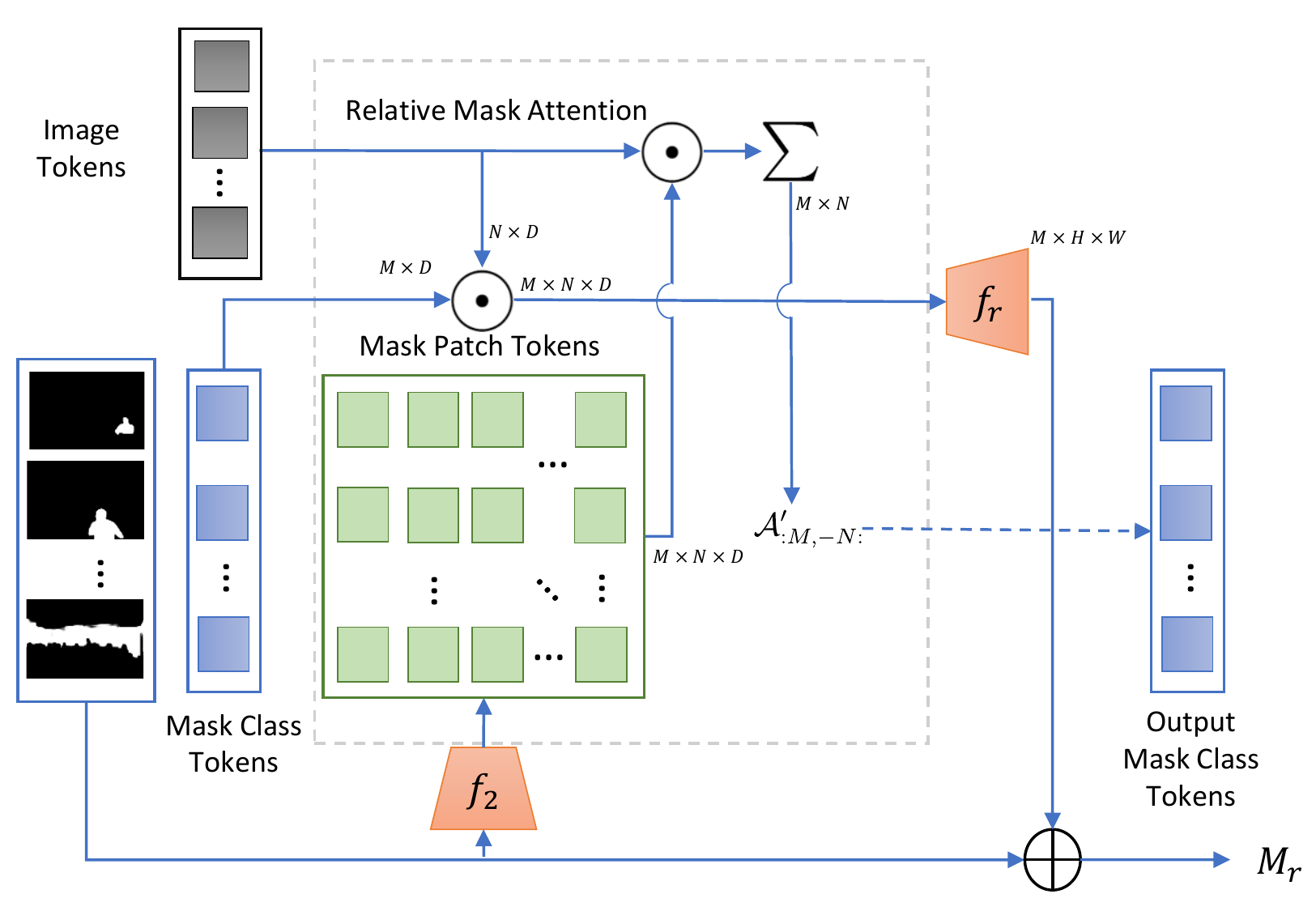}
  \caption{\small \textbf{Relative Mask Attention.} Our Relative Mask Attention mechanism adds another attention matrix $A'_{:M, -N:}$ to the original attention matrix. The newly added attention matrix is computed using the Image Tokens and the Mask Patch Tokens. The mask patch tokens are acquired by patchifying the masks using a similar way for the images as shown here. Moreover, the masks are refined by using $M_r$ in Eq. \ref{eq:refine} which is computed by Image Tokens and Mask Class Tokens.} 
\label{fig:rma}
\end{center}
 \end{figure}
\paragraph{Relative Mask Attention.} To further utilize the mask information and refine the coarse masks, we propose Relative Mask Attention mechanism in our transformer. Our key design principle is to try not to change the CLIP features directly as this would destroy the learned relationship between the image features and text features in the CLIP model. Therefore, we adopt a way to only change the attention matrix in the transformer to learn a better linear combination of the values in the attention layers according to the mask information. As in Figure \ref{fig:rma}, our proposed Relative Mask Attention Mechanism only changes the attention matrix and refines the masks. $M_r$ is defined in Eq. \ref{eq:refine}. $\calA^{'}_{:M, -N:}$ is defined in Eq. \ref{eq:a1}. $f_M$ is the class-agnostic mask proposal network. $f_1$ and $f_2$ are two downsampling networks that encode the images/masks to image tokens/mask patch tokens sharing the same architecture. $f_r$ is a two-layer convolutional network that maps the attention matrix to a mask residual.

Similar to relative positional encoding, we use a relative attention mechanism here. Let $D$ be the dimension of the token embedding, 
for each Mask Class Token $T^{\text{MC}}_i\in \mathbb{R}^{D}$ with a corresponding mask $K_i\in \mathbb{R}^{H\times W}$ whose shape is the same as the image, we use a similar way as for the images to get mask patch tokens $T^{\text{MP}}\in \mathbb{R}^{M\times N\times D}$ in the computation of the attention. In our attention matrix, the Mask Class Tokens attending the image tokens part will then be as follows:
\begin{align}
    \label{eq:a1}
    \calA^{'}_{:M, -N:} &= \sum_{c}^{D}(\phi_{Q_m}(T^{\text{MP}})\odot \phi_{K_m}(T^{\text{IM}}))_{c} \\
    \calA_{:M, -N:} & = \frac{\phi_Q(T^{\text{MC}})\cdot \phi_K(T^{\text{IM}}) + \calA^{'}_{:M, -N:}}{2\sqrt{D}}
\end{align}
where $T^{\text{IM}}\in \mathbb{R}^{N\times D}$ is image tokens, $T^{\text{MC}}\in \mathbb{R}^{M\times D}$ is Mask Class Tokens, $T^{\text{MP}}\in \mathbb{R}^{M\times N\times D}$ is Mask Patch Tokens $\phi_Q, \phi_K, \phi_{Q_m}, \phi_{K_m}$ are linear transformations, $\odot$ is element-wise product and $\sum_c^{D}(\cdot)_c$ is the sum of the embedding dimension. $\phi_{K_m}(T^{\text{IM}}) \in \mathbb{R}^{N\times D}$ will first be broadcast to $\mathbb{R}^{M\times N\times D}$ before doing element-wise production.

The attention will also in turn be used for the refinement of the masks. The vanilla attention can be seen as a relationship between each mask area and all the image patches. Thus we utilize this to help our coarse masks be more accurate. The updating process of the masks is as following:
\begin{align}
\label{eq:refine}
    M_r = \sigmoid(\sigmoid^{-1}(M_c) + f_r(\phi_Q(T^{\text{MC}})\odot \phi_K(T^{\text{IM}})))
\end{align}
where $M_c, M_r\in \mathbb{R}^{N\times H\times W}$ denotes the coarse mask and refined mask respectively, $f_r$ is a learnable non-linear function that maps the attention matrix to a mask residual, $\sigmoid$ and $\sigmoid^{-1}$ are sigmoid and inverse sigmoid functions respectively.

The RMA method aims to leverage detailed mask information and refine masks by utilizing CLIP's features. Without RMA, the method would only utilize the mask information in the attention mask (which is just a low-resolution mask) and cannot refine the mask using CLIP's features. In order to utilize the detailed information of masks, we add another attention matrix, which is obtained from the Mask Patch Tokens and the Image Tokens, to the original attention matrix in the CLIP ViT model so that the new attention matrix could be aware of the detailed mask information and thus the Mask Class Tokens could attend the information more accurately. Furthermore, we use the information from the original attention matrix, which is obtained from the Mask Class Tokens and the Image Tokens, to refine the mask.

\section{Experiments}

In this part, we train our proposed MaskCLIP method using COCO  \citep{lin2014microsoft} training data and test on other datasets (ADE20K  \citep{zhou2016semantic, zhou2017scene}, PASCAL Context  \citep{mottaghi_cvpr14}, LVIS) under the open vocabulary setting. We report our results on semantic/instance/panoptic segmentation tasks to evaluate the performance of out model's universal segmentation.
\subsection{Datasets}

\textbf{COCO:} COCO \citep{lin2014microsoft} includes 133 classes where 80 classes are things and 53 classes are stuff or background. There are 118k training images and 5k validation images. In our experiments, we first train the class-agnostic mask proposal network on COCO training dataset using the annotations of panoptic masks. Then we train our models on COCO training images in a supervised manner.

\textbf{ADE20K:} ADE20K  \citep{zhou2016semantic, zhou2017scene} contains 20,210 images and annotations for training and 2000 images and annotations for validation. It serves both panoptic segmentation and semantic segmentation. The full version (A-847)  \citep{zhou2016semantic} includes 847 classes and the short version (A-150)  \citep{zhou2017scene} includes 150 classes. We use the validation set in ADE20K for testing without any training on this dataset in which case we can test our model's capability of open vocabulary segmentation. 

\textbf{PASCAL Context:} PASCAL Context  \citep{mottaghi_cvpr14} contains 10,103 per-pixel annotations for images of PASCAL VOC 2010  \citep{pascal-voc-2010}, where 4998 for training and 5105 for validation. The full version (P-459) includes 459 classes and the short version includes 59 classes. This dataset serves as another benchmark testing our model's open vocabulary segmentation ability.

\textbf{LVIS:} LVIS \citep{gupta2019lvis} contains 100,170 images for training and 19,809 images for validation. It extends COCO \citep{lin2014microsoft} but contains 1,203 categories. It is considered one of the most challenging benchmark for instance segmentation because of its large vocabulary, long-tailed distribution, and fine-grained classification. We report our model's performance of open vocabulary instance segmentation on the validation dataset. 

\vspace{-3mm}
\begin{table*}[!htp]
\centering
\caption{\small \textbf{Results on open-vocabulary semantic segmentation.}  A-150 and A-847 represent the ADE20K dataset with 150 classes and 847 classes respectively. P-459 and P-59 represent PASCAL Context dataset with 459 classes and 59 classes respectively. All results use the mIoU metric. All methods presented here don't use extra data other than COCO for training. }
\label{tab:semantic_results}
\vspace{-1mm}
\scalebox{0.8}{
\begin{tabular}{l | l | c | c | c | c}

\textbf{Method } & COCO Training Data & A-150 $\uparrow$ & A-847 $\uparrow$ & P-459 $\uparrow$ & P-59 $\uparrow$\\
\hline
ALIGN \citep{jia2021scaling} & None & 10.7 & 4.1 & 3.7 & 15.7 \\
ALIGN w/ proposals  \citep{jia2021scaling} & Masks & 12.9 & 5.8 & 4.8 & 22.4 \\
LSeg+ \citep{li2022language} & Masks + Labels & 18.0 & 3.8 & 7.8 & 46.5 \\
OpenSeg \citep{ghiasi2021open}
          & Masks + Captions & 21.1 & 6.3 & 9.0 & 42.1 \\
SimSeg \citep{xu2022simple} & Masks + Labels & 20.5 & 7.0 & - & \textbf{47.7} \\
\hline
CLIP Baseline & Masks & 13.8 & 5.2 & 5.2 & 25.3\\
MaskCLIP w/o RMA & Masks & 14.9 & 5.6 & 5.3 & 26.1 \\
MaskCLIP (MaskRCNN) & Masks + Labels & 22.4 & 6.8 & 9.1 & 41.3 \\
MaskCLIP
          & Masks + Labels & \textbf{23.7} & \textbf{8.2} & \textbf{10.0} & 45.9
\end{tabular}
}
\end{table*}

\begin{figure*}[!htp]
\begin{center}
\scalebox{0.8}{
\begin{tabular}{ccccc}

\includegraphics[width=0.16\textwidth]{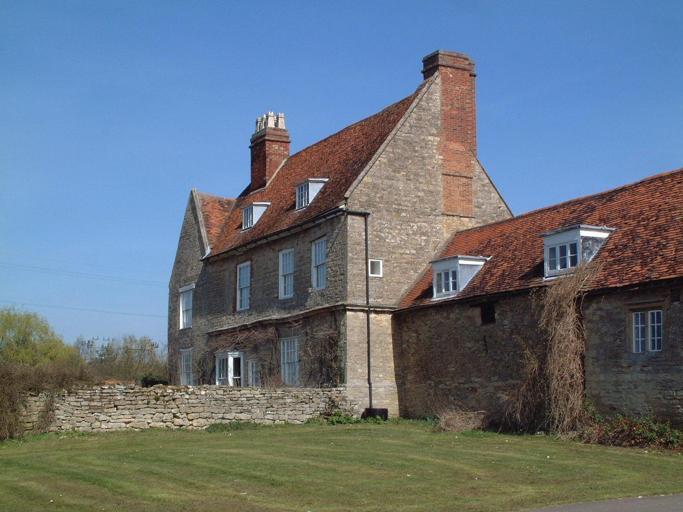} & \hspace{-4mm}
\includegraphics[width=0.16\textwidth]{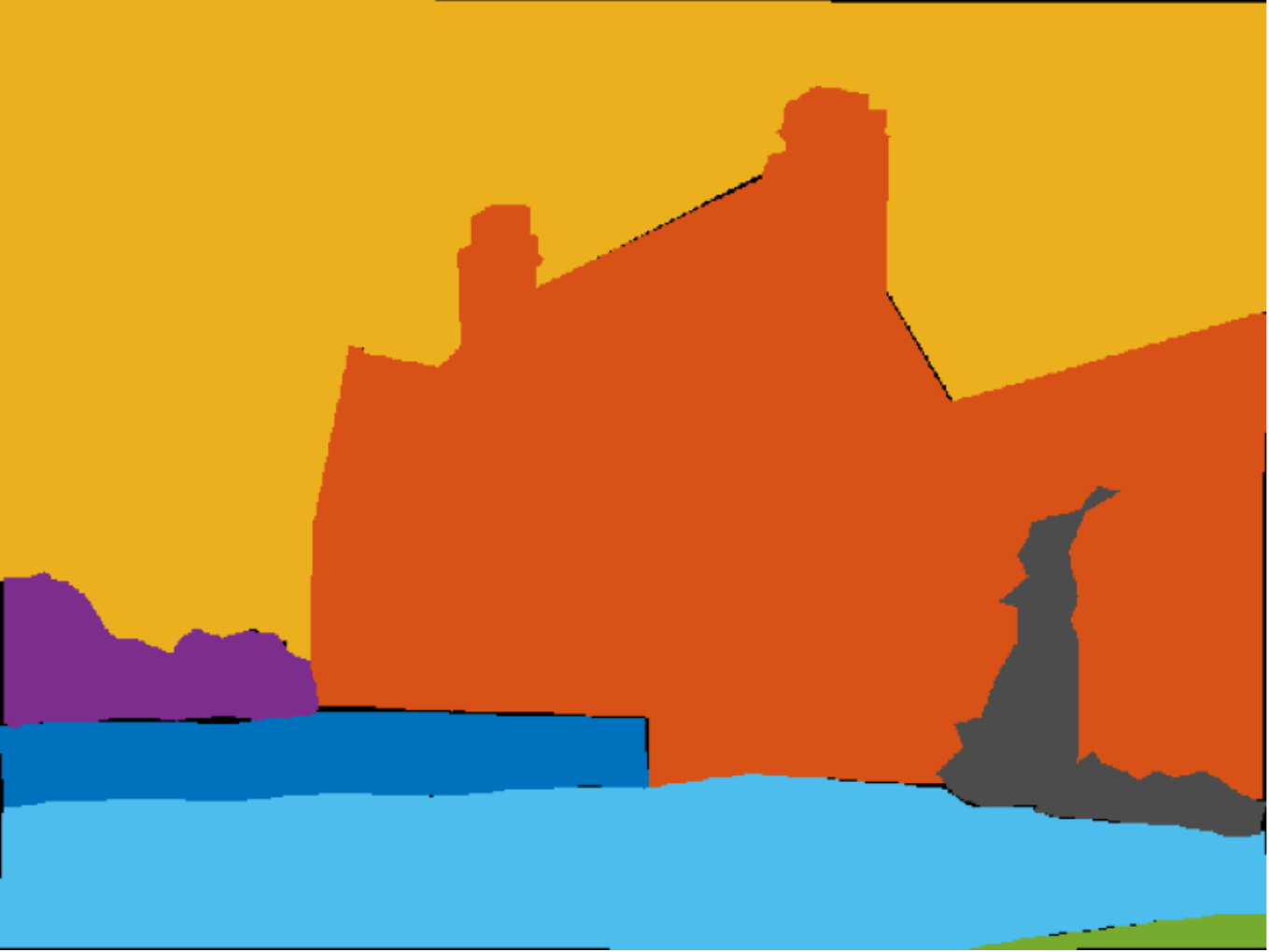} &
\hspace{-4mm}
\includegraphics[width=0.16\textwidth]{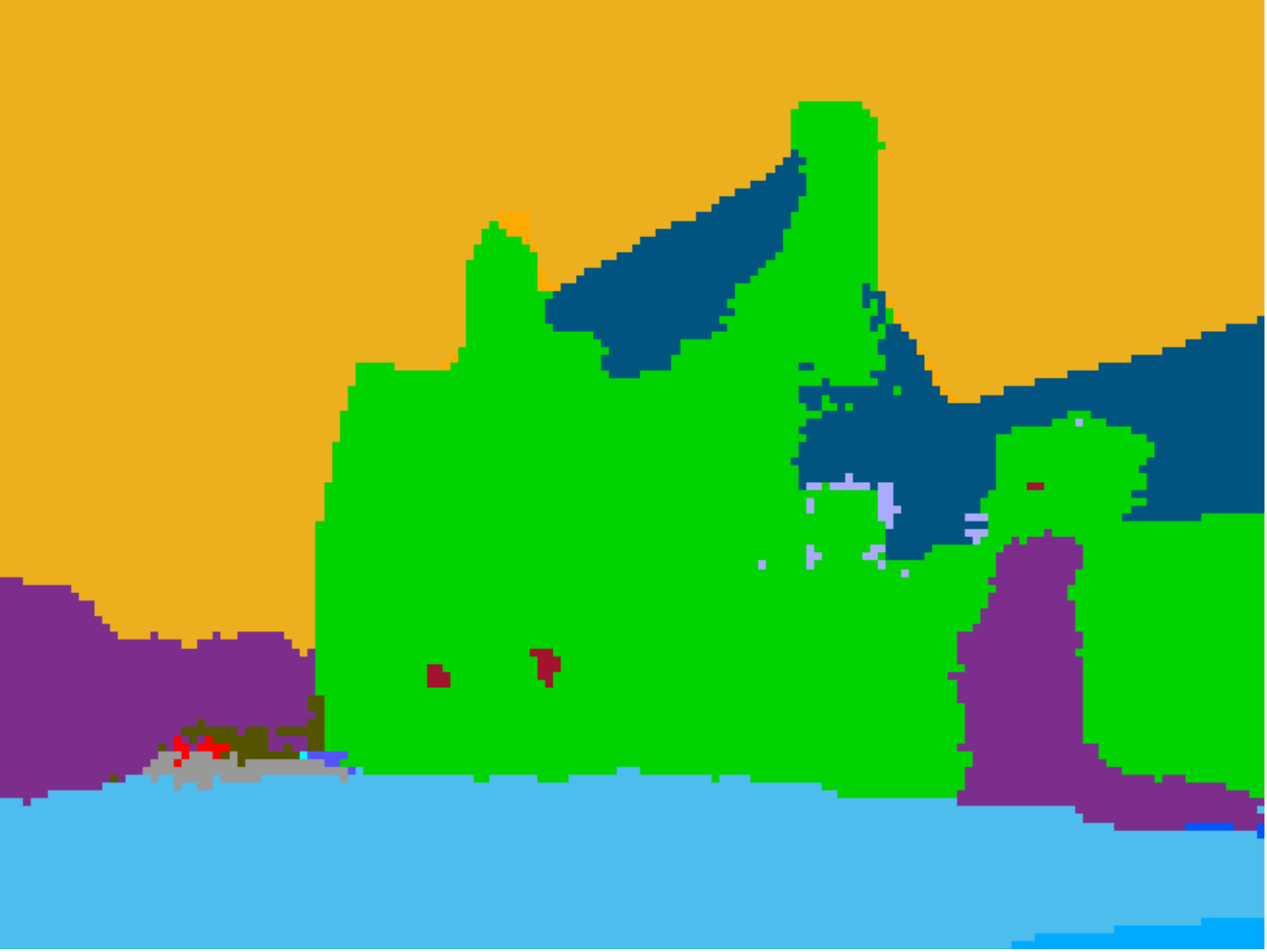} & 
\hspace{-4mm}
\includegraphics[width=0.16\textwidth]{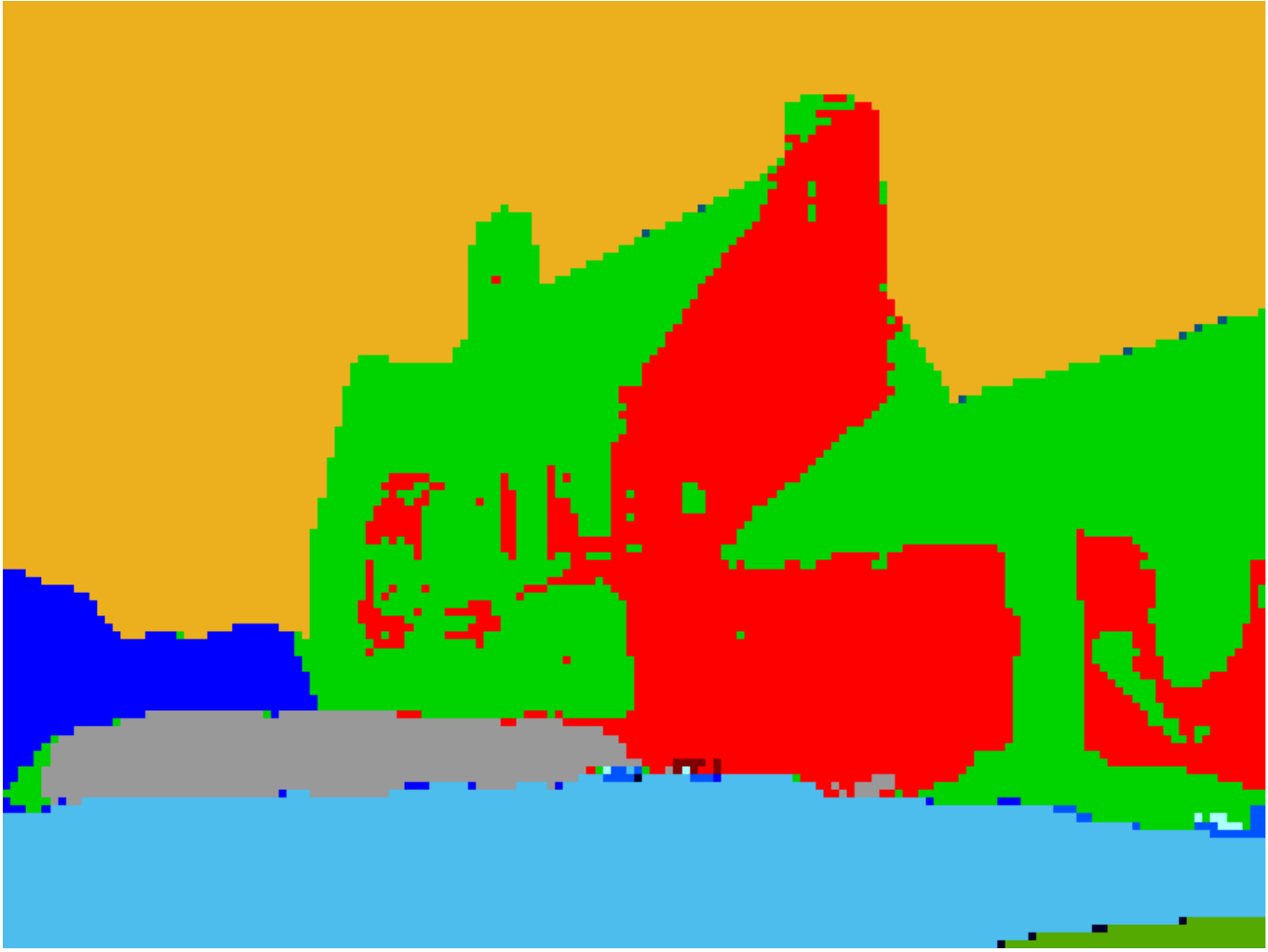} &
\hspace{-4mm}
\includegraphics[width=0.16\textwidth]{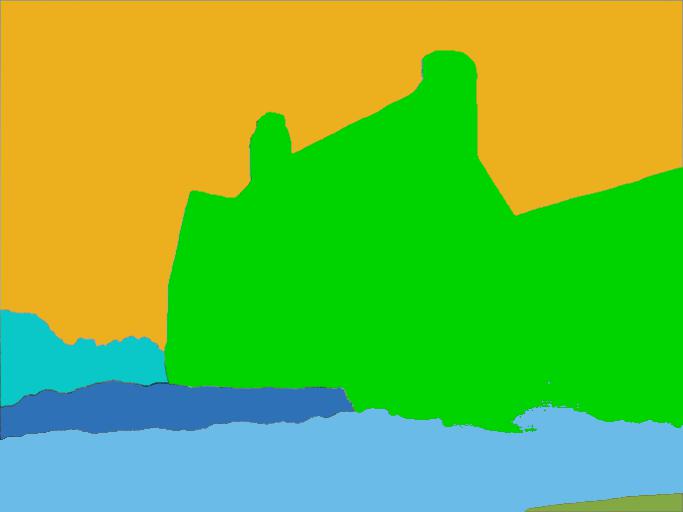} 
\\
Image & \hspace{-4mm} GT & \hspace{-4mm} ALIGN++ & \hspace{-4mm} OpenSeg & \hspace{-4mm} MaskCLIP \\
\multicolumn{5}{c}{\includegraphics[width=0.7\textwidth]{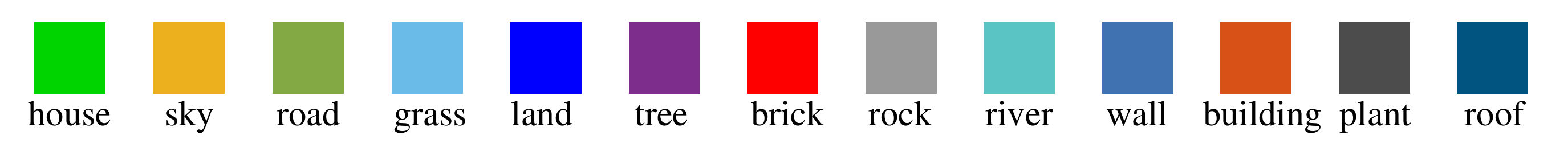}}

\end{tabular}}
\caption{\small Comparison on open-vocabulary semantic segmentation. The input image and the results for GT, ALIGN++, OpenSeg are from  \citep{ghiasi2021open}.}

\label{fig:comp}
\end{center}
\end{figure*}

\subsection{Implementation Details}

\textbf{Class-Agnostic Mask Proposal Network.} In our first stage, we train a class-agnostic mask proposal network using MaskRCNN \citep{he2017mask} and Mask2Former \citep{cheng2021masked} on COCO training data. The experiment setting we use for MaskRCNN is R50-FPN-1x. The backbone we use in Mask2Former is ResNet-50. All the training setting follows the default in their models.

\textbf{CLIP Baseline.} We design our first baseline by directly using the class-agnostic mask proposal network from the first stage and the pretrained CLIP model. We mask the images according to the masks from the class-agnostic mask proposal network and send the masked images to the CLIP model to get classification results. The pretrained CLIP model we use is ViT-L/14@336px and the text inputs we use are simply the category names defined by each dataset. Those two settings keep the same with the following two methods for a fair comparison.

\textbf{MaskCLIP w/o RMA Baseline.} Our second baseline is based on the Mask Class Tokens which doesn't use the Relative Mask Attention mechanism. Instead of masking the images and sending the resulting images directly to the CLIP model for feature extraction, we use Mask Class Tokens to acquire the corresponding partial/dense image features. The obtained image features will then be used for further open vocabulary classification.

The two baselines above don't need any training in the second stage and can be used to directly perform the open vocabulary tasks. We will demonstrate that the second baseline is better at feature extraction in both quantitative results and qualitative results under the open vocabulary setting and show the effectiveness and efficiency of the proposed Mask Class Tokens.

\textbf{MaskCLIP.} In our MaskCLIP method, we still use the CLIP ViT-L/14@336px pretrained model as with the previous two. This model has 24 attention layers and we add Relative Mask Attention in four of them which is 6, 12, 18, 24. We use AdamW \citep{loshchilov2017decoupled} as our optimizer and the learning rate is set to 0.0001. We train our model on COCO training data for 10k iterations with a batch size of 8. The training takes around 3h on 8 Nvidia A5000 GPUs.

{
\textbf{Loss Function.} The loss function is $\mathcal{L} = \lambda_{\text{ce}} \mathcal{L}_{\text{ce}} + \lambda_{\text{dice}} \mathcal{L}_{\text{dice}} + \lambda_{\text{bce}} \mathcal{L}_{\text{bce}}$, where $ \mathcal{L}_{\text{ce}}$ is the loss for classification, $\mathcal{L}_{\text{dice}}$ and  $\mathcal{L}_{\text{bce}}$ are the losses for mask localization. In our experiments, We set $\lambda_{\text{ce}}=2, \lambda_\text{dice}=5, \lambda_\text{bce}=5$. 
}

In the next three parts, we evaluate our methods on open vocabulary semantic, panoptic segmentation, and instance segmentation tasks. The class-agnostic mask proposal networks we use in those methods are trained using Mask2Former other than noted.

\subsection{Open-Vocabulary Semantic Segmentation}

First, we use our method to compare with open-vocabulary semantic segmentation as in Table \ref{tab:semantic_results}. We train our method on the COCO dataset and evaluate on another four different datasets. On the four datasets we test, MaskCLIP outperforms the two baselines we described in the implementation details which demonstrates that our feature extraction method is better than the vanilla way in this setting. It extracts the features without the need to change the input and can simultaneously extract multiple mask area features easily. For 100 masks' feature extraction in a single image, the CLIP baseline takes \textasciitilde 3s on a single 3090 GPU while the MaskCLIP w/o RMA baseline only takes \textasciitilde 0.6s which is \textbf{\textasciitilde 4x} faster. Our MaskCLIP beats both baselines significantly as it utilizes accurate mask information and refines the masks during the feature extraction process. Furthermore, our proposed method also reaches state-of-the-art results on three of the benchmarks with only P-59 slightly lower than LSeg+\cite{li2022language}.

To compare with previous methods, we also provide a semantic segmentation comparison in Figure \ref{fig:comp}. Results on ALIGN++ and OpenSeg are directly from  \citep{ghiasi2021open} and we run the same image using our MaskCLIP model. It can be seen that due to the open vocabulary setting, some similar classes may be mistakenly classified e.g. all three methods predict the house in this image while the ground truth is building.

\begin{table*}[!htbp]
\begin{center}
\caption{\small \textbf{Results on open-vocabulary panoptic segmentation using the ADE20k validation dataset.} th and st represent thing and stuff classes respectively. }
\label{tab:ade20k_val_results}
\scalebox{1.0}{
\begin{tabular}{l | ccc | ccc | ccc}

Method  & PQ $\uparrow$ & PQ$^{\text{th}}$ $\uparrow$ & PQ$^{\text{st}}$ $\uparrow$ &  SQ $\uparrow$ & SQ$^{\text{th}}$ $\uparrow$ & SQ$^{\text{st}}$ $\uparrow$  &  RQ $\uparrow$ & RQ$^{\text{th}}$ $\uparrow$ & RQ$^{\text{st}}$ $\uparrow$ \\
\hline

CLIP Baseline & 8.207 & 8.473 & 7.675 & 53.124 & 52.661 & 54.048 & 10.534 & 10.883 & 9.835 \\
MaskCLIP w/o RMA &
           9.565 & 8.922 & 10.852 & 62.507 & 62.268 & 62.985 & 12.645 & 11.758 & 14.418 \\
\hline
MaskCLIP (MaskRCNN) & 12.860 & 11.242 & 16.095 & 64.008 & 64.183 & 63.658 & 16.803 & 14.968 & 20.473 \\
MaskCLIP & 
           \textbf{15.121} & \textbf{13.536} & \textbf{18.290} & \textbf{70.479} & \textbf{70.021} & \textbf{71.396} & \textbf{19.211} & \textbf{17.448} & \textbf{22.737} \\
\end{tabular}
}
\end{center}
\end{table*}

\subsection{Open-Vocabulary Panoptic Segmentation}

\label{sec:quantitative_panoptic}

Next, we compare our MaskCLIP with the two baselines on ADE20K validation set under the open vocabulary panoptic segmentation setting. The results are presented in Table  \ref{tab:ade20k_val_results}. As can be seen from the table, the MaskCLIP w/o RMA baseline performs better on all the metrics in panoptic segmentation setting which further demonstrates our method's effectiveness.

\begin{figure*}[!hpt]
\centering

    \scalebox{1.0}{
\begin{tabular}{cc@{\hskip 6pt}c@{\hskip 6pt}c}

    \rotatebox[origin=c]{90}{Input Image} &
    \raisebox{-.5\height}{\includegraphics[height=1.24in]{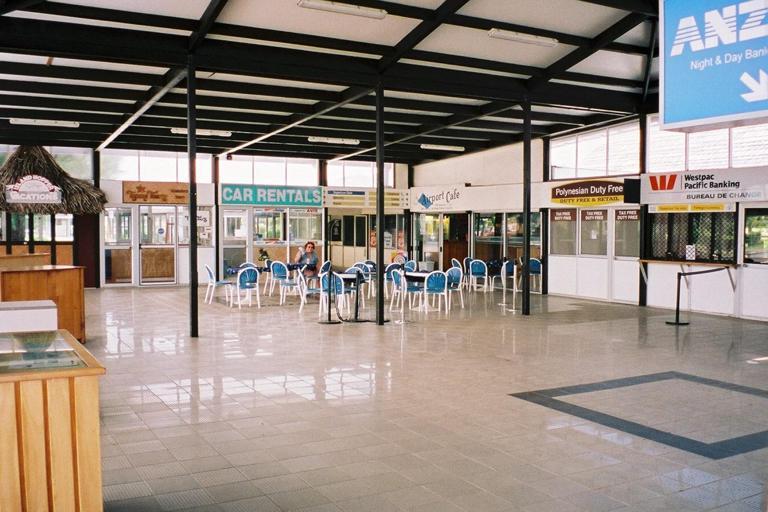}} & 
    \raisebox{-.5\height}{\includegraphics[height=1.24in]{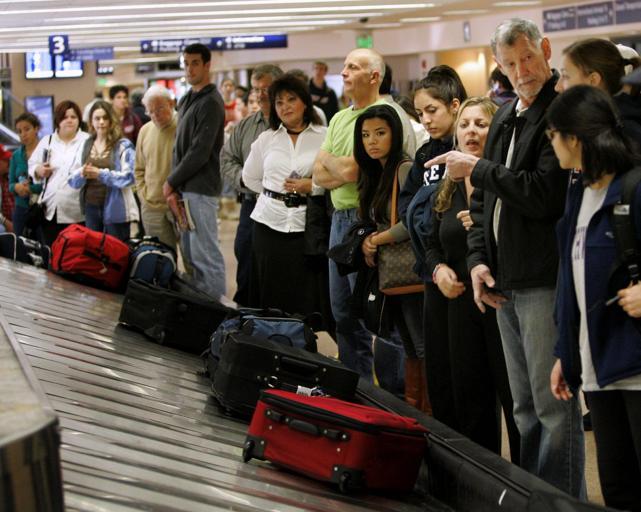}} & 
    \raisebox{-.5\height}{\includegraphics[height=1.24in]{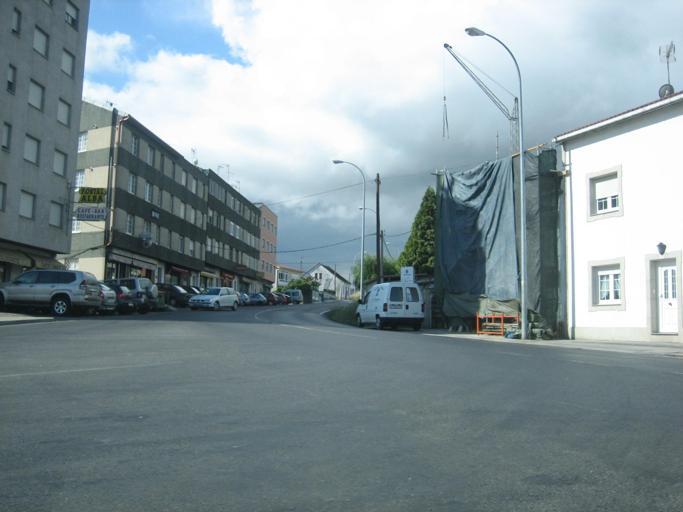}} \vspace{3pt}\\
    
    \rotatebox[origin=c]{90}{CLIP Baseline} &
    \raisebox{-.5\height}{\includegraphics[height=1.24in,angle=0]{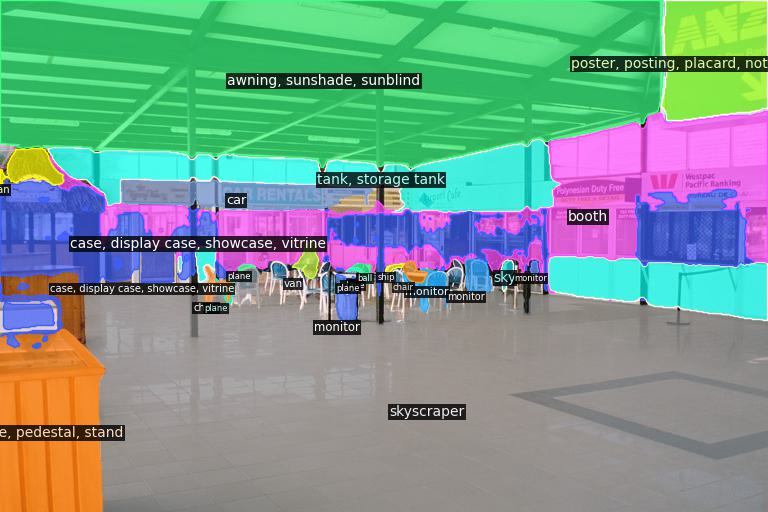}} & 
    \raisebox{-.5\height}{\includegraphics[height=1.24in,angle=0]{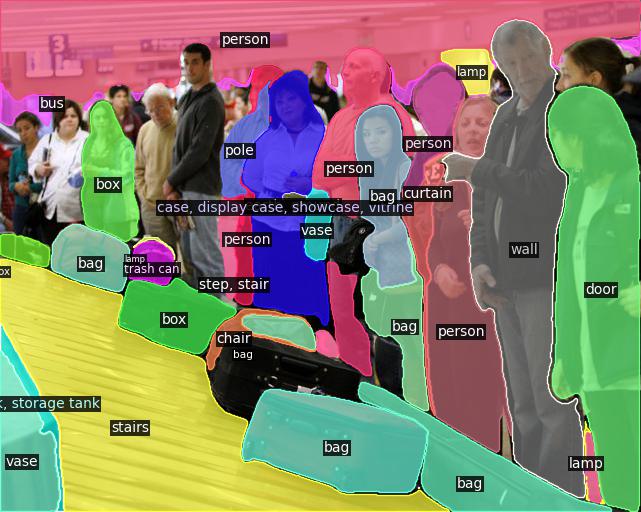}} & 
    \raisebox{-.5\height}{\includegraphics[height=1.24in,angle=0]{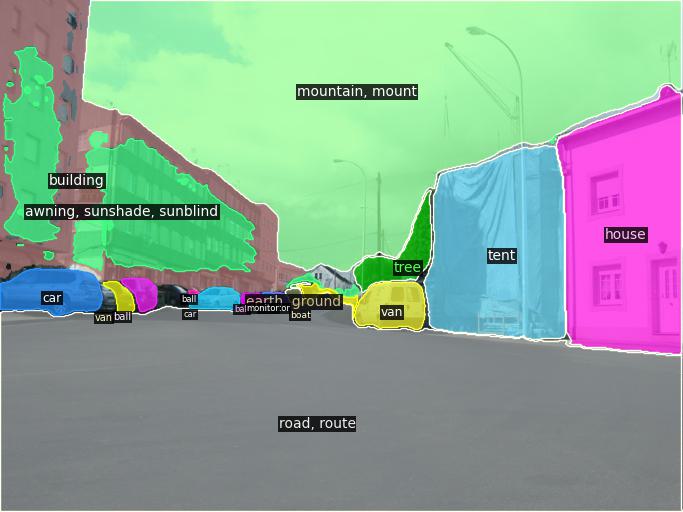}} \vspace{3pt}\\
    
    \rotatebox[origin=c]{90}{MaskCLIP w/o RMA} &
    \raisebox{-.5\height}{\includegraphics[height=1.24in,angle=0]{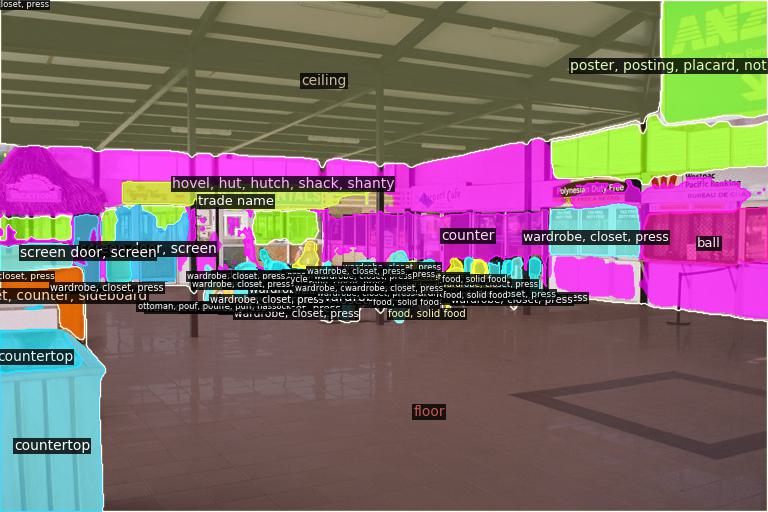}} & 
    \raisebox{-.5\height}{\includegraphics[height=1.24in,angle=0]{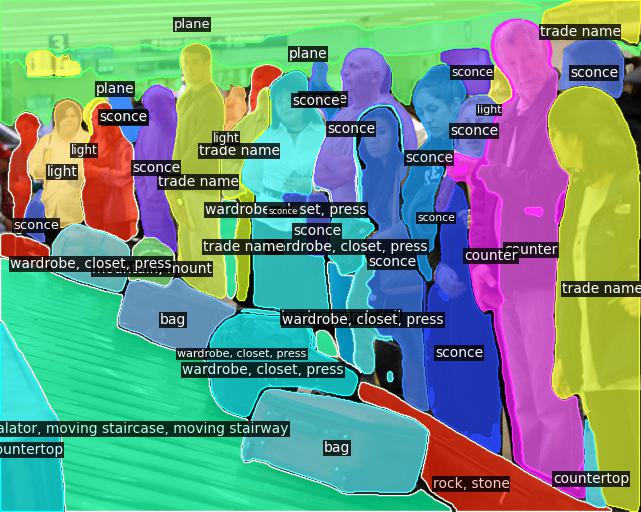}} & 
    \raisebox{-.5\height}{\includegraphics[height=1.24in,angle=0]{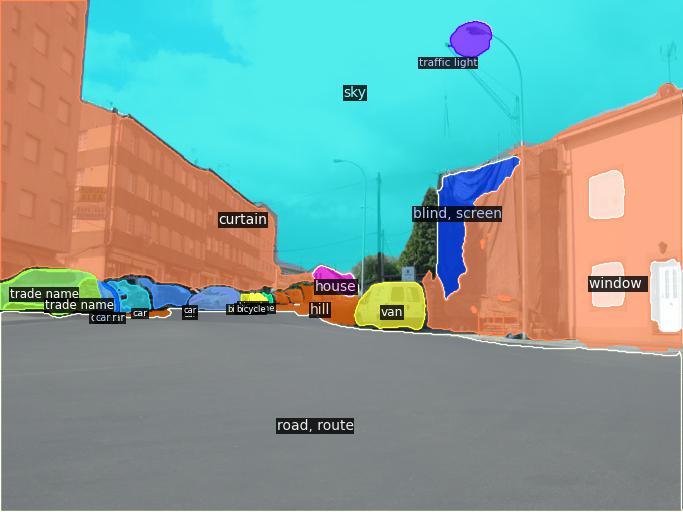}} \vspace{3pt} \\
    
    \rotatebox[origin=c]{90}{MaskCLIP} &
    \raisebox{-.5\height}{\includegraphics[height=1.24in,angle=0]{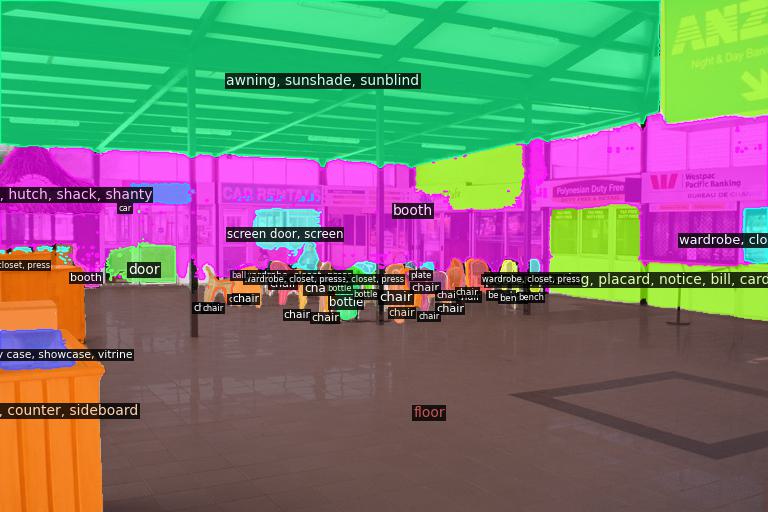}} & 
    \raisebox{-.5\height}{\includegraphics[height=1.24in,angle=0]{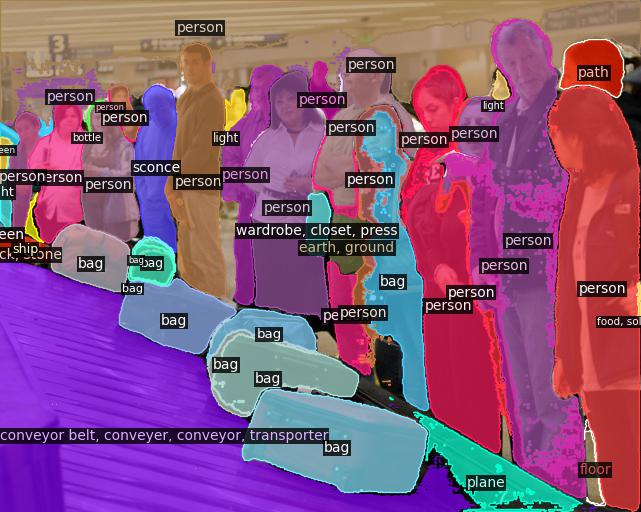}} & 
    \raisebox{-.5\height}{\includegraphics[height=1.24in,angle=0]{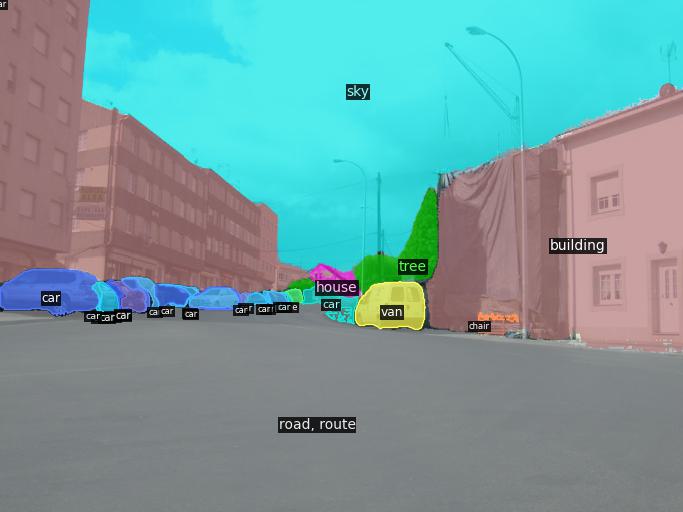}}
\end{tabular}}

\caption{\small \textbf{Qualitative results on ADE20K panoptic segmentation.} The images are taken from the ADE20K validation set. We use the class names directly from the ADE20K 150 classes as the text inpputs. Three images are presented here using our MaskCLIP model along with the two baselines.}

\label{fig:panoptic}
\vspace{-1mm}
\end{figure*}

We also show two sets of images to demonstrate our model capability. The first is the qualitative results on ADE20K. We compare our method with the two baselines in Figure \ref{fig:panoptic}. It can be seen that our method performs much better than the two baselines. The results from the first column show that due to the lack of global information, CLIP baseline fails to predict ``floor''. Instead, it predicts ``skyscraper''. 
While the MaskCLIP w/o RMA baseline and MaskCLIP model can predict the floor correctly as it does not lose the global context information.

\begin{figure*}[!htp]
\begin{center}
\scalebox{0.95}{
\begin{tabular}{ccc}

\includegraphics[height=0.2\linewidth]{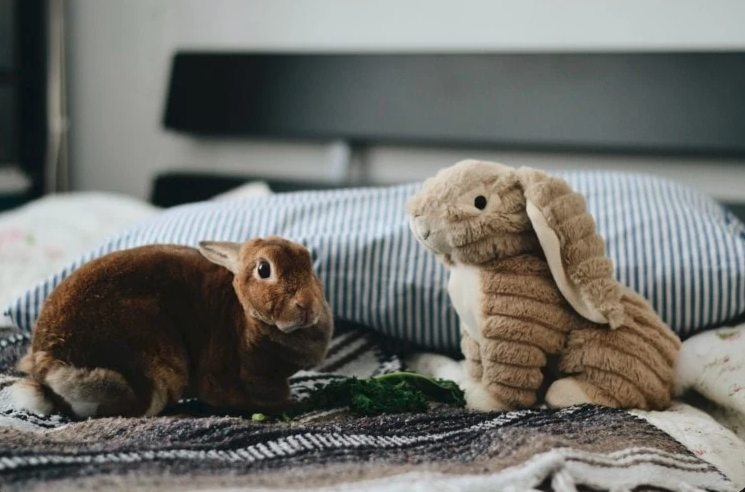} & \hspace{-2mm}
\includegraphics[height=0.2\linewidth]{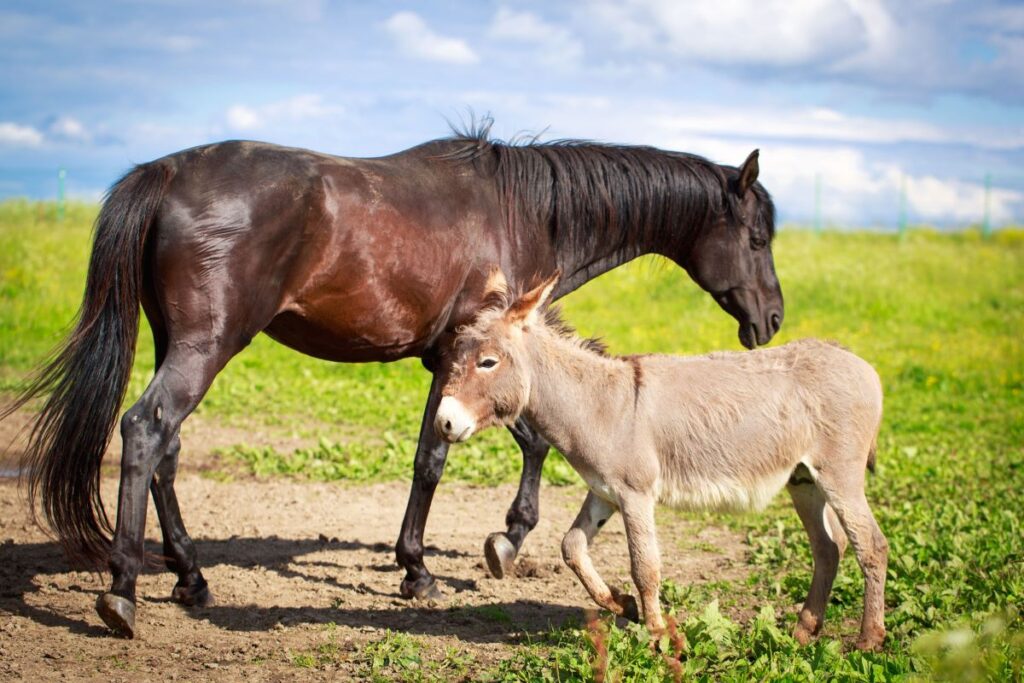} & \hspace{-5mm}
\includegraphics[height=0.2\linewidth]{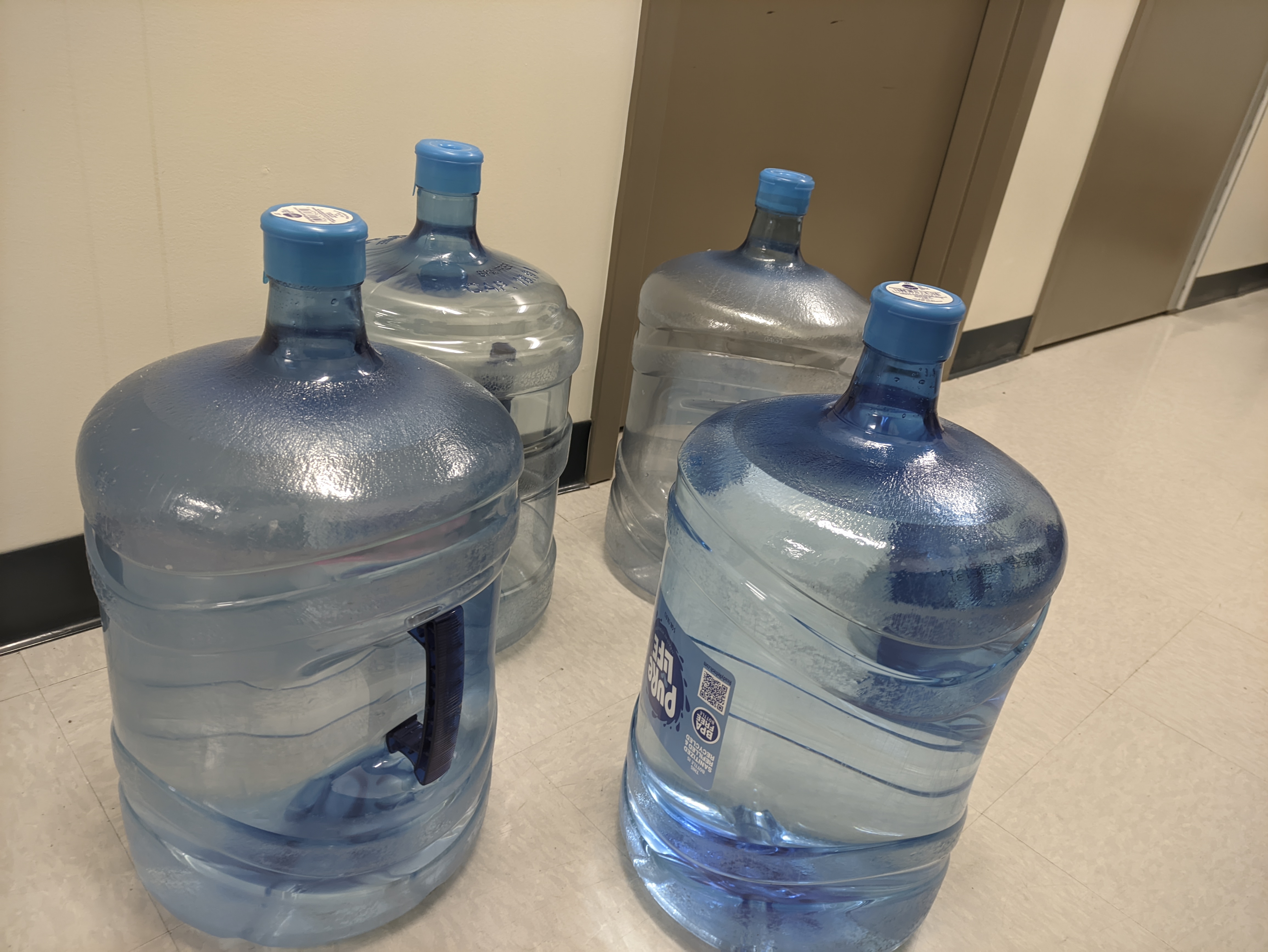}
\\

\includegraphics[height=0.2\linewidth]{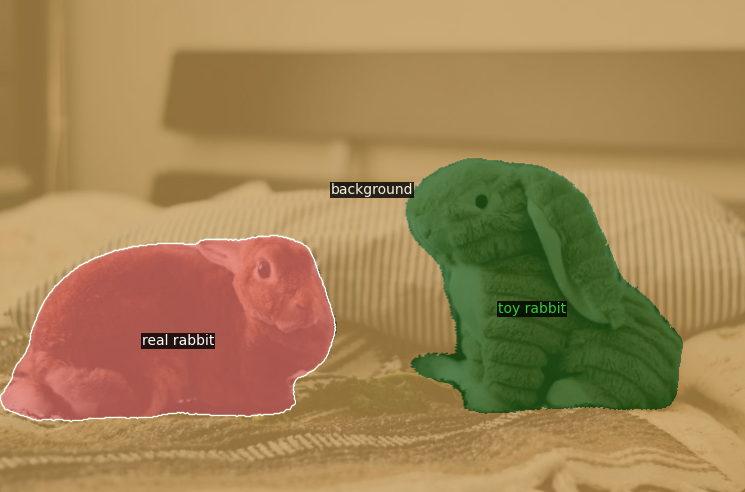} &  \hspace{-2mm}
\includegraphics[height=0.2\linewidth]{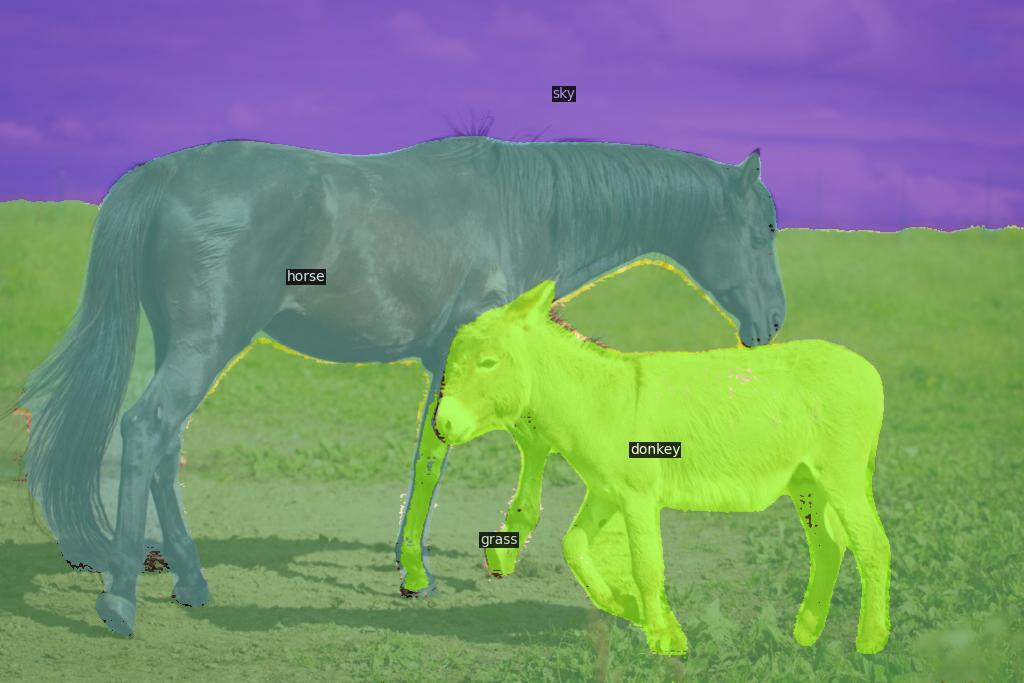} &\hspace{-5mm}
\includegraphics[height=0.2\linewidth]{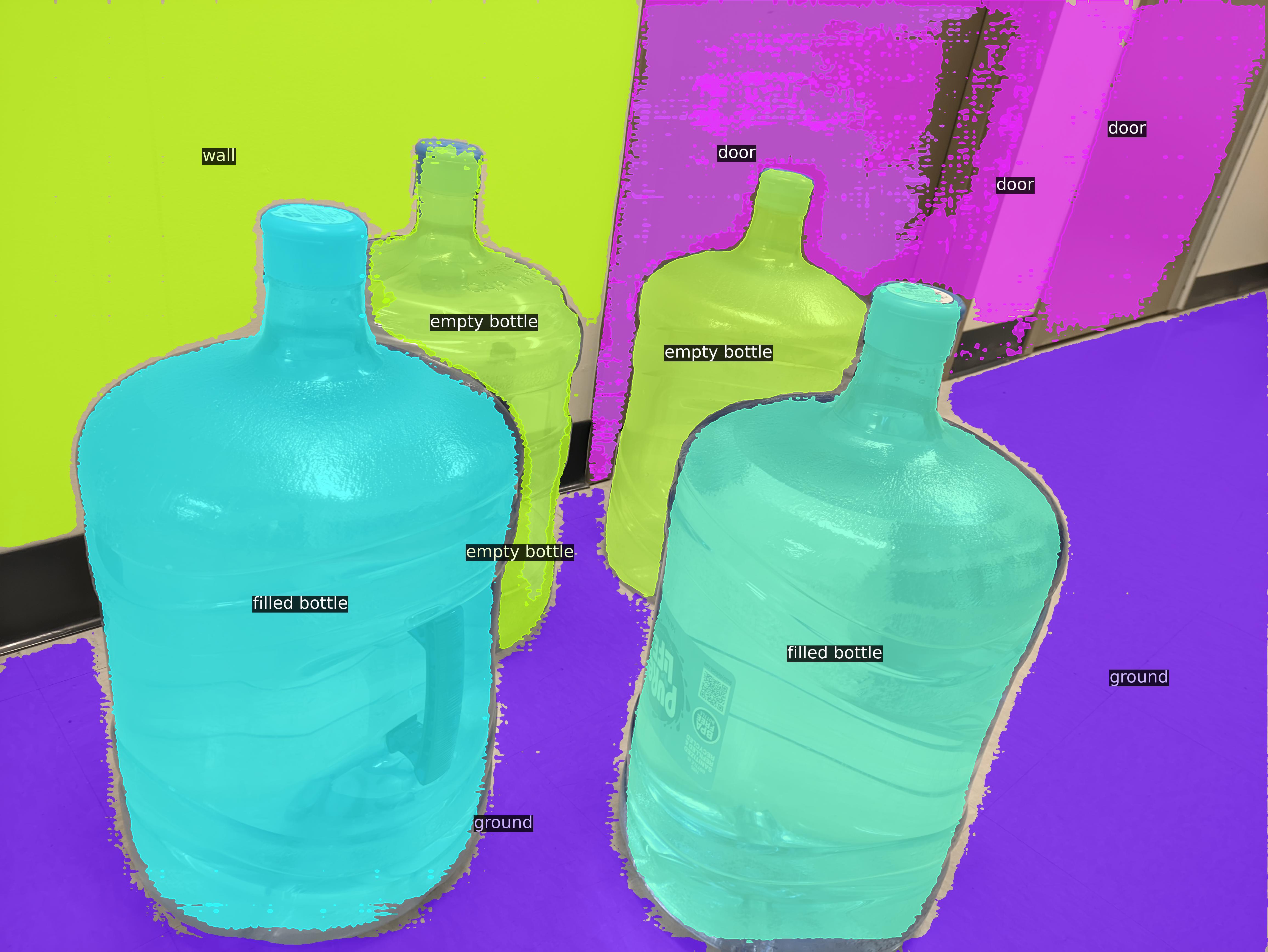}
\\

(a) {\scriptsize ``\textbf{toy rabbit}'', ``\textbf{real rabbit}'',}  & (b) {\scriptsize ``horse'', ``\textbf{donkey}'',}  & (c) {\scriptsize ``\textbf{empty bottle}'', ``\textbf{filled bottle}'',}  \\
{\scriptsize ``\textbf{background}''} & {\scriptsize ``sky'', ``grass''} & {\scriptsize ``door'', ``wall'', ``ground''}
\end{tabular}
}

\caption{\small \textbf{User-specified class panoptic segmentation.} The labels above are the text inputs we used for testing the images. Texts in bold are novel classes i.e. don't exist in the labels of COCO training data. (a) Our model is able to distinguish object properties of real rabbit and toy rabbit. (b) This example shows that our model is potential for fine-grained classifications and does not have bias toward the base classes. (c) Our results show that it can tell the difference between the filled status and empty status of bottles.}
\label{fig:novel}
\end{center}
\end{figure*}

The second set of images we're presenting is in Figure \ref{fig:novel}. These figures show our capability of specifying any arbitrary classes in performing panoptic segmentation task. The results show that though we train a new model based on the CLIP model without any distillation methods, we can still preserve the CLIP image features very well. Our model doesn't have a clear bias towards the base classes in the training set and could tell the difference very well that have no chance to learn in the COCO training: e.g toy vs real and filled vs empty.

\subsection{Open-Vocabulary Instance Segmentation}

\paragraph{Cross-Dataset Setting.} We present the results on open vocabulary instance segmentation in Table \ref{tab:ade20k_ins_val_results} under the cross-dataset setting. Since instance segmentation can be regarded as ``thing-only`` panoptic segmentation, we directly apply our model trained on COCO panoptic dataset to the instance segmentation task. MaskCLIP with different class-agnostic mask proposal networks performs better than CLIP Baseline and MaskCLIP w/o RMA in general. 

\begin{table}[H]
\begin{center}
\caption{\small \textbf{Results on open-vocabulary instance segmentation under the cross-dataset setting.} }
\label{tab:ade20k_ins_val_results}
\scalebox{0.7}{
\begin{tabular}{l | c c c | c c c}
\multirow{2}{*}{Method} & \multicolumn{3}{c|}{ADE20K} & \multicolumn{3}{c}{LVIS} \\
  & AP $\uparrow$ & AP$^{\text{50}}$ $\uparrow$ & AP$^{\text{75}}$ $\uparrow$ & AP $\uparrow$  & AP$^{\text{50}}$ $\uparrow$ &  AP$^{\text{75}}$ $\uparrow$  \\
\hline

CLIP Baseline & 3.974 & 6.090 & 4.288 & 4.989 & 7.244 & 5.227
            \\
MaskCLIP w/o RMA & 4.263 & 6.696 & 4.402 &  5.762 & 8.202 & 6.169
            \\
\hline
MaskCLIP (MaskRCNN) & \textbf{6.164} & \textbf{12.072} & 5.775 & 6.431 & \textbf{12.753} & 5.777 \\
MaskCLIP & 5.989 & 9.739 & \textbf{6.209}  & \textbf{8.404} & 12.190 & \textbf{8.810}\\
\end{tabular}
}
\end{center}
\end{table}

\paragraph{COCO Split Setting.} Besides the cross-dataset setting, we also follow the COCO Split Setting in XPM\citep{huynh2022open} to perform the instance segmentation in Table \ref{tab:cocosplit}. On the generalized setting which is a more challenging setting, we outperform previous results in base, target, and all categories. In the constrained setting, we also show competitive results in both base and target categories.

\begin{table}[H]
\begin{center}
\caption{ \small \textbf{Results on open-vocabulary instance segmentation under the COCO split setting.}}
\label{tab:cocosplit}
\scalebox{0.75}{
\begin{tabular}{>{}l | >{}c >{}c | >{}c >{}c >{}c}
\multirow{2}{*}{Method} & \multicolumn{2}{c|}{ Constrained} & \multicolumn{3}{c}{  Generalized} \\
  & Base & Target & Base & Target & All \\
\hline

Soft-Teacher\citep{xu2021end} &	41.8&	14.8&	41.5&	9.6	&33.2 \\ 
Unbiased-Teacher\citep{liu2021unbiased} &	41.8&	15.1&	41.4&	9.8&	33.1 \\
XPM\citep{huynh2022open} &	42.4&	\textbf{24.0}&	41.5&	21.6&	36.3 \\
\hline
MaskCLIP & \textbf{42.8}	&23.2&	\textbf{42.6}&	\textbf{21.7}&	\textbf{37.2}\\
\end{tabular}
}
\end{center}
\end{table}

\subsection{Efficiency Analysis}

We further provide efficiency analysis in Table \ref{tab:flops} to demonstrate the efficiency of our feature extraction method. Previous methods usually perform a crop/mask operation on the input images and send the resulted images to CLIP to obtain the partial/dense features for classification which is rather slow. In contrast, our proposed method employs Mask Class Tokens to obtain the partial/dense features for classification. By doing so, our method can extract partial/dense features more efficiently (instead of running CLIP $N$ times, our method only requires running CLIP one time with $N$ more Mask Class Tokens) and is also aware of the global context information.

\begin{table}[H]
\centering
\caption{\textbf{FLOPs Comparison.} We use the CLIP ViT-L/14 model in all methods for fair comparison and 640x640 as the input resolution.}
\label{tab:flops}
\scalebox{1.0}{
\begin{tabular}{l | r }
  Method & TFLOPs \\
  \hline
 RegionCLIP\cite{zhong2021regionclip} & 9.5 \\
 ZegFormer\cite{ding2021decoupling} & 10.3 \\
 SimSeg\cite{xu2022simple} & 9.6 \\
 CLIP Baseline & 10.5  \\
 \hline
 MaskCLIP(Ours) & \textbf{0.3}
\end{tabular}
}
\end{table}

\section{Ablation Study} 

\subsection{Incorporating GT Masks.}

Since our model can decouple the mask proposal process and the classification process, we could also use the ground truth mask proposals which can be regarded as a ``perfect'' mask proposal network in our method. In this way, we can eliminate the effects of the quality of the mask proposals and inspect the method's classification capabilities.  In Table \ref{tab:ablation_gtmasks}. We can see that the performance could gain a lot from the ``perfect'' mask proposals. And our MaskCLIP method also outperforms OpenSeg in this setting.

\begin{table}[H]
\centering
\caption{\textbf{Incorporating GT Masks.} Results on using GT masks as mask proposals for open-vocabulary panoptic segmentation and semantic segmentation.}
\label{tab:ablation_gtmasks}
\scalebox{0.9}{
\begin{tabular}{l | c | c }
  & PQ $\uparrow$ & mIoU $\uparrow$ \\
  \hline
OpenSeg \citep{ghiasi2021open} & - & 21.1 \\
MaskCLIP & 15.1 & 23.7 \\
\hline
OpenSeg + GT masks \citep{ghiasi2021open} & - & 27.5 \\
MaskCLIP + GT masks & 35.8 & 31.7  \\
\end{tabular}
}
\end{table}

\subsection{Mask Refinement.}

In our Relative Mask Attention part, the attention layer will use the accurate mask information to learn a better attention matrix and the mask will also use the attention information to gradually refine itself. In this ablation study, we only let the attention matrix learn from the mask without any mask refinement. And we get the results in Table \ref{tab:ablation_mr}. Since the SQ reflects the segmentation quality, we care more about SQ here. It can be seen that MaskCLIP performs slightly better than that without the mask refinement which demonstrates the effectiveness of the mask refinement.

\begin{table}[!htp]
\begin{center}
\caption{\textbf{Ablation Study on Mask Refinement.} Results on ADE20K validation set are reported here. Both methods are trained on COCO and tested on ADE20K validation dataset. MR refers to mask refinement.}
\label{tab:ablation_mr}
\scalebox{0.71}{
\begin{tabular}{l | c | c | c | c | c | c}
  & PQ $\uparrow$ & PQ$^{\text{Th}}$ $\uparrow$ & PQ$^{\text{St}}$ $\uparrow$ & \textbf{SQ}$\uparrow$ & \textbf{SQ$^{\text{Th}}$} $\uparrow$ & \textbf{SQ$^{\text{St}}$} $\uparrow$\\
  \hline
MaskCLIP w/o MR & 13.624 & 13.253 & 14.368 & 66.361 & 67.715 & 63.653\\
MaskCLIP & 15.121 & 13.536 & 18.290 & 70.479 & 70.021 & 71.396 \\
\end{tabular}
}
\end{center}
\end{table}

\section{Conclusion}

In this paper, we have presented a new algorithm, MaskCLIP, to tackle an emerging computer vision task, open-vocabulary universal image segmentation. MaskCLIP is a Transformer-based approach using mask queries with the ViT-based CLIP backbone to efficiently and effectively utilize pre-trained partial/dense CLIP features. MaskCLIP consists of a Relative Mask Attention (RMA) module that is seamlessly integrated with a pre-trained CLIP. MaskCLIP is distinct  compared with prior approaches in open-vocabulary semantic segmentation/object detection by building an integrated encoder module for segmentation mask refinement and image feature extraction with a pre-trained CLIP image model. Encouraging experimental results on open-vocabulary semantic/instance/panoptic segmentation have been obtained.

\indent {\bf Acknowledgement} This work is supported by NSF Award IIS-2127544. We thank Xiang Zhang and Boyi Li for helpful discussions.

\bibliography{example_paper}
\bibliographystyle{icml2023}


\newpage
\appendix
\onecolumn


\section{CLIP Baseline Details}

\begin{figure}[H]
\centering
\scalebox{0.8}{
\includegraphics[height=0.3\linewidth]{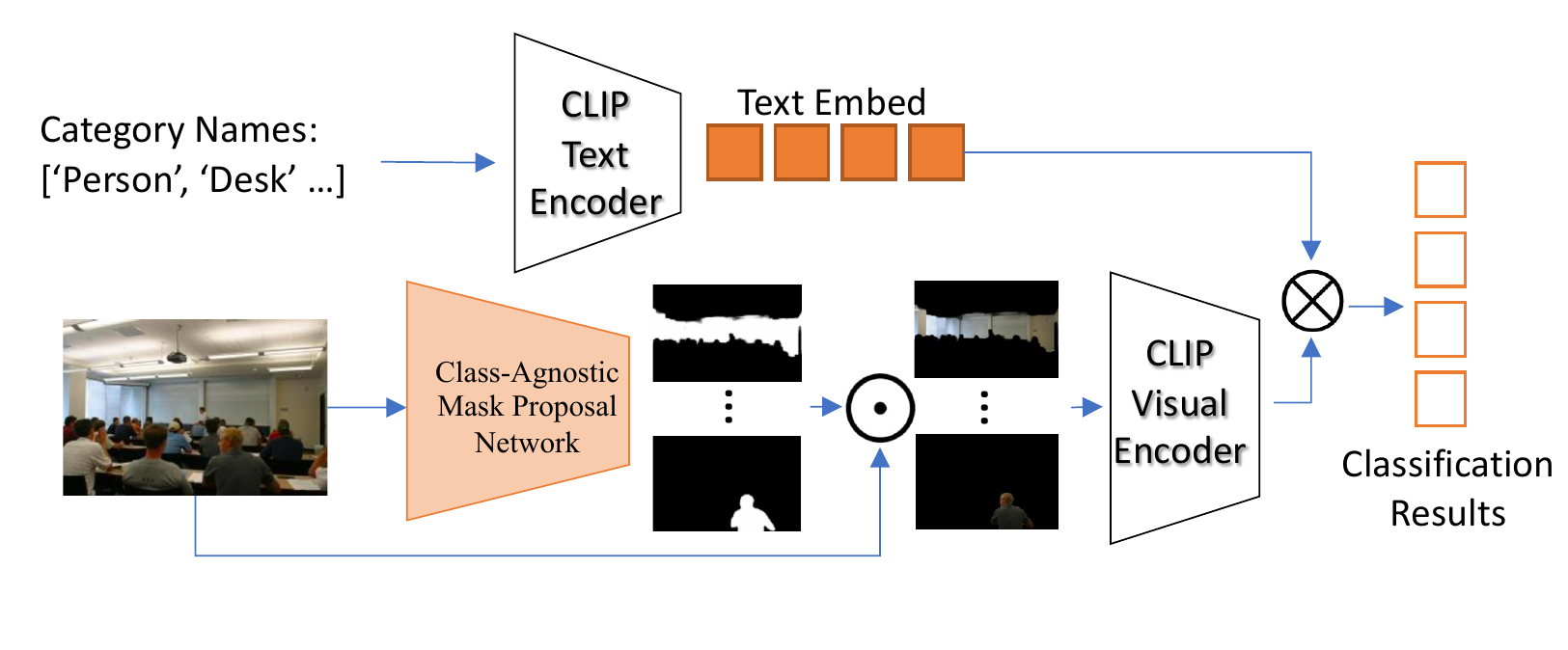}
}
\caption{Illustration of the CLIP baseline.}
\label{fig:illus-baseline}
\end{figure}

Here we provide more details on our CLIP Baseline. Given an RGB image $\mathcal I \in \mathbb{R}^{H\times W \times 3}$ with height $H$ and width $W$ and a list of category names with $C$ classes, we precompute the text embedding of the category names as $\mathcal E \in \mathbb{R}^{C\times D}$. The mask proposal network $f_m$ outputs $N$ masks $\mathcal M \in \mathbb{R}^{N \times H \times W}$. For each mask: the cropped image region is the element-wise product between the binary mask $\mathcal M_i$ and the image $\mathcal I$, i.e. $\mathcal R_i \in \mathbb{R}^{H\times W\times 3}$; the visual embedding $\mathcal V_i \in \mathbb{R}^D$ of the cropped region is computed by the visual encoder where $D$ is the hidden dimension; the final classification score $\mathcal Y_i \in \mathbb{R}^C$ is the softmax over the dot product between the visual embedding $\mathcal V_i$ and the text embedding $\mathcal T$. 
A formal algorithm is described as \ref{alg:maskclip-baseline} and a visualization of this is shown as \ref{fig:illus-baseline}.

\begin{algorithm}[h]
	\caption{CLIP Baseline}
	\begin{algorithmic}
	    \Require Mask proposal network $f_m$, CLIP visual encoder $f_v$, CLIP text encoder $f_t$.
		\State Given an image $\mathcal{I} \in \mathbb{R}^{H\times W \times 3}$ and a list $\mathcal T$ containing $C$ category names.
		\State $\mathcal E = f_t(\mathcal T)$.
		\State $\mathcal M = f_m(\mathcal I)$.
		\For{$t = 1,2,\dots,N $}
		\State $\mathcal R_i = \mathcal M_i \odot \mathcal I$.
		\State $\mathcal V_i = f_v(\mathcal R_i)$.
		\State $\mathcal Y_i = \operatorname{softmax}(\mathcal E \otimes \mathcal V_i)$.
		\EndFor
	\end{algorithmic}
	\label{alg:maskclip-baseline}
\end{algorithm}

\section{Ablation on using Relative Mask Attention in Different Layers} 

In this part we conduct an ablation study on using different layers for relative mask attention. Since our pretrained CLIP model is fixed during the whole training procedure, whether each layer would help the final results remains a question. We use four different kinds of layers combination of the layers in this part and provide the results in Table \ref{tab:ablation_rma}. We can see that the last layer is a key part of our results since the features are gradually learned through all the attention layer. Though the last four layers' features should the best, the performance wouldn't be better if Relative Mask Attention is only used in the last four layers. This is also reasonable since the network should not have the accurate mask information too late.

\begin{table}[!htbp]
\begin{center}
\caption{\textbf{Ablation Study on Relative Mask Attention Layers in different layers.} All the methods are trained on COCO and tested on ADE20K validation dataset. The pretrained CLIP ViT-L/14@336px model has 24 layers and we replace four of them with our relative mask attention to fully utilize the accurate mask information and refine the masks. }

\label{tab:ablation_rma}
\begin{tabular}{l | c | c | c}
 Different Layers & PQ & PQ$^{\text{Th}}$ & PQ$^{\text{St}}$\\
\hline
1, 7, 13, 19 & 11.241 & 10.519  & 12.686 \\
3, 9, 15, 21 & 11.372 & 10.141 & 13.835 \\
21, 22, 23, 24 & 14.673 & 14.048 & 15.922 \\
6, 12, 18, 24 & 15.121 & 13.536 & 18.290
\end{tabular}
\end{center}
\end{table}

\section{More Visualization Results on Arbitrary Categories}

\begin{figure*}[!htbp]
\begin{center}
\begin{tabular}{ccc}

\includegraphics[height=0.2\linewidth]{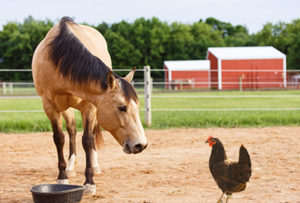} & \hspace{-2mm}
\includegraphics[height=0.2\linewidth]{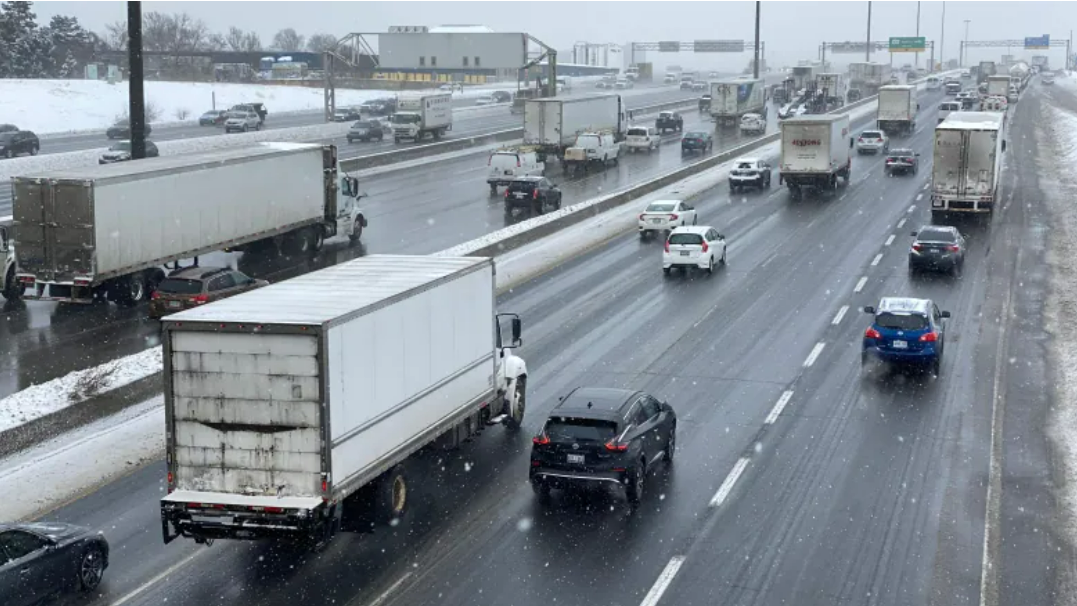} & \hspace{-5mm}
\includegraphics[height=0.2\linewidth]{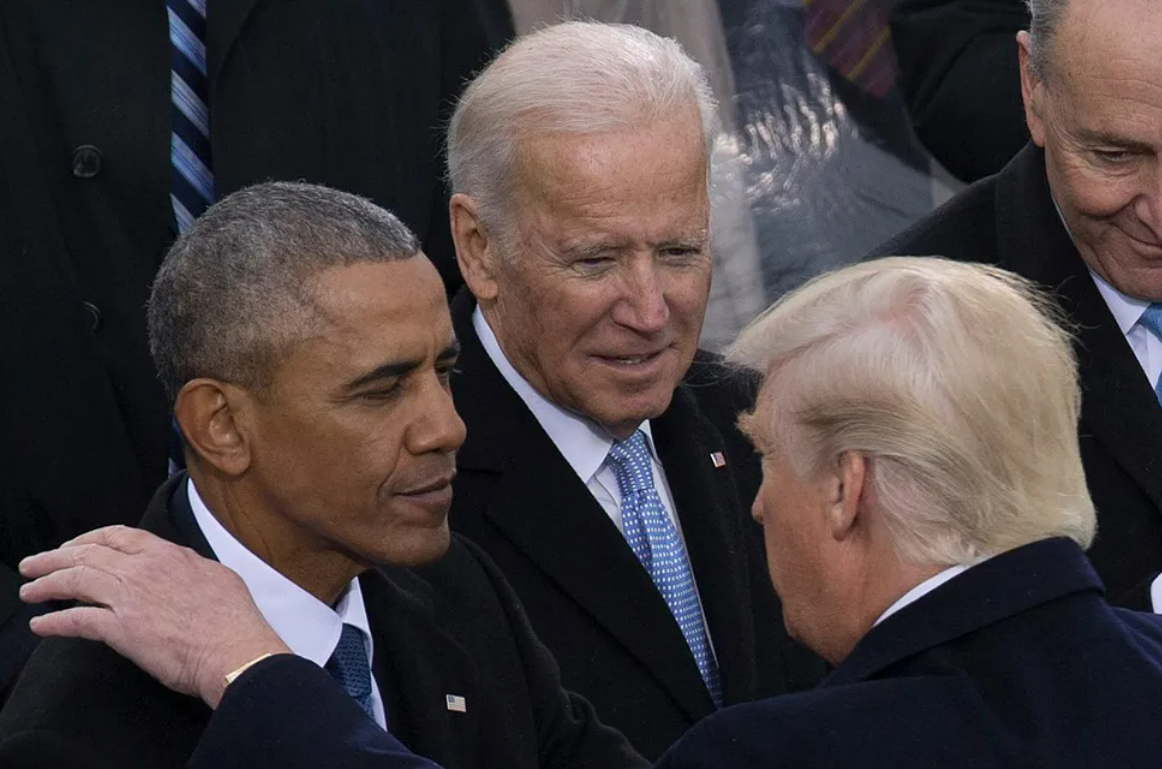}
\\

\includegraphics[height=0.2\linewidth]{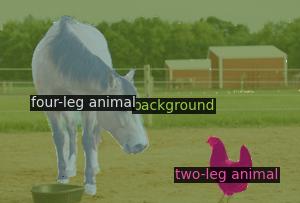} &  \hspace{-2mm}
\includegraphics[height=0.2\linewidth]{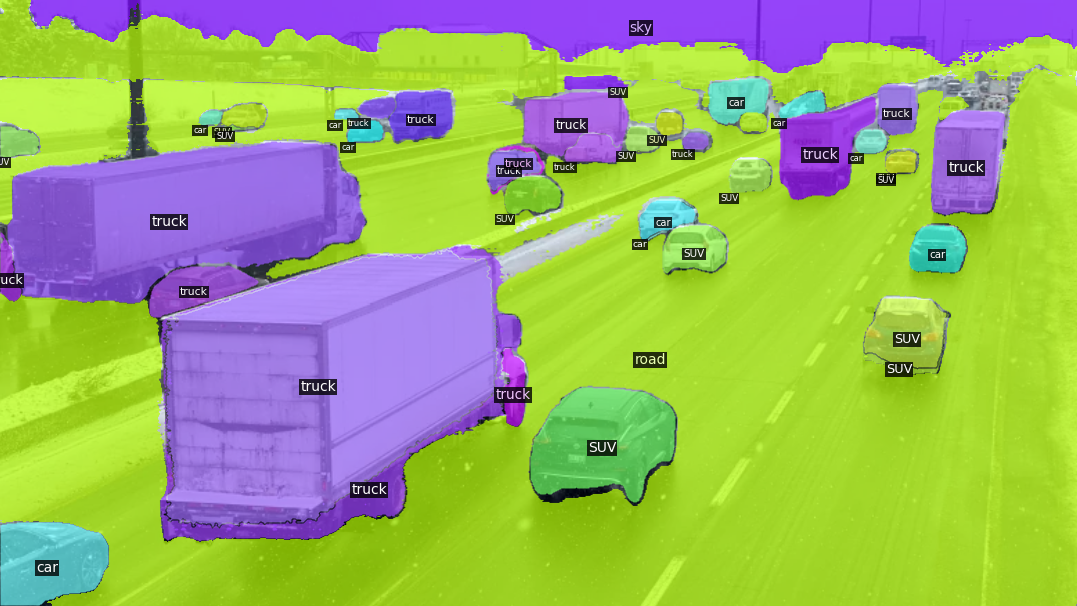} &\hspace{-5mm}
\includegraphics[height=0.2\linewidth]{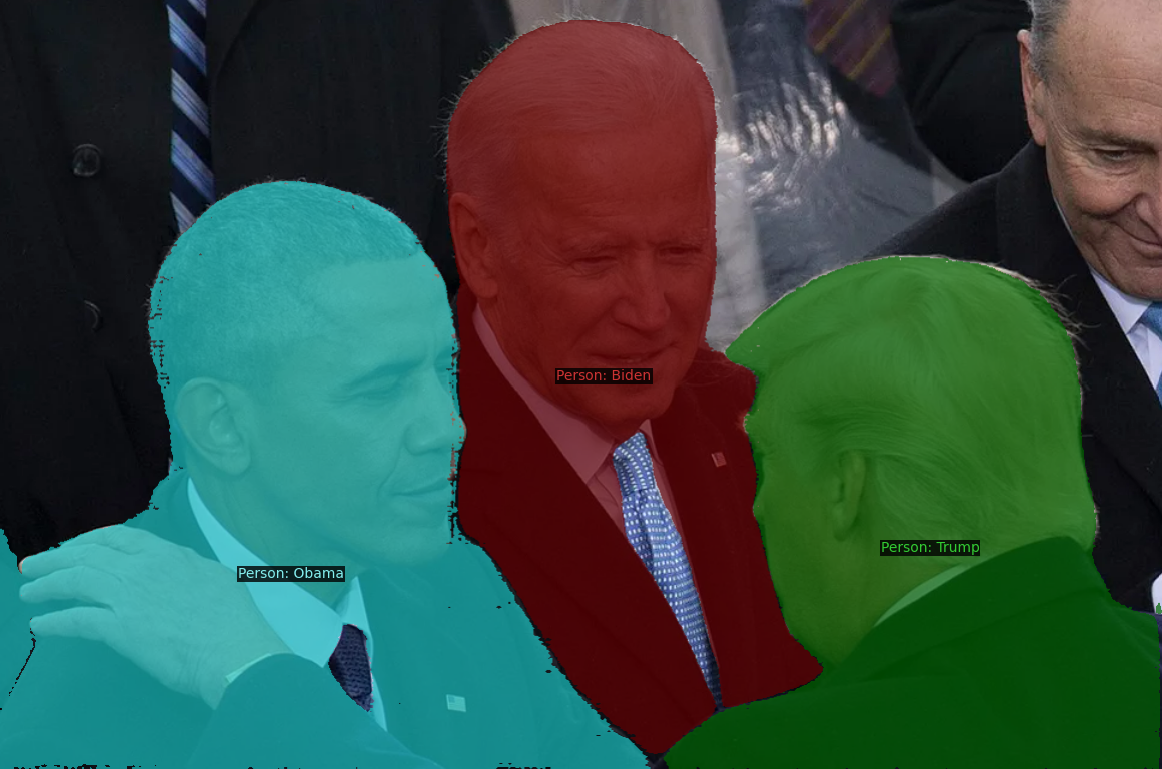}
\\

(a) {\scriptsize ``\textbf{four-leg animal}'', ``\textbf{two-leg animal}'',}  & (b) {\scriptsize ``car'', ``truck'', ``\textbf{SUV}''}  & (c) {\scriptsize ``\textbf{Person: Obama}'', ``\textbf{Person: Biden}'',}  \\
{\scriptsize ``\textbf{background}''} & {\scriptsize ``road'', ``sky''} & {\scriptsize ``\textbf{Person: Trump}''}
\end{tabular}

\caption{\small \textbf{More qualitative restuls on user-specified class.} The labels above are the text prompts we used for testing the images. Texts in bold are novel classes i.e. don't exist in the labels of COCO training data.}
\label{fig:app_novel}
\end{center}
\end{figure*}

In this part, we provide more visualization results on user-specified class discoveries in Figure \ref{fig:app_novel}. We select some very close text prompts such as ``four-leg animal" and ``two-leg animal"; ``car'', ``truck'' and ``SUV'' and find that our method can still classify them. We also show another result which is ``person identification'' in Figure \ref{fig:app_novel} (c) which shows our model preserve the dense/local CLIP features rather well.

\end{document}